\documentclass[a4paper]{styles/svproc}  
\usepackage{url}

\usepackage{bigints}

\usepackage{epsfig} 
\usepackage{comment} 
\usepackage{mathptmx} 
\usepackage{times} 
\usepackage{amsmath} 
\usepackage{amssymb} 
\usepackage{caption} 
\usepackage{subfigure}
\usepackage{ascmac}
\usepackage{epsf}
\usepackage{bm}
\usepackage[percent]{overpic}
\usepackage{multicol}
\usepackage{hyperref}
\graphicspath{{./fig/all}}

\renewcommand{\bf}{\bfseries}

\newcommand{\eq}[1]{(\ref{eq:#1})}
\newcommand{\fig}[1]{Fig.\ \ref{fig:#1}}
\newcommand{\tab}[1]{Table\ \ref{tab:#1}}

	\newcommand{\parallelsum}{\mathbin{\!/\mkern-5mu/\!}}
	
\renewcommand{\baselinestretch}{0.97} 

\let\llncssubparagraph\subparagraph
\let\subparagraph\paragraph
\usepackage[compact]{titlesec}
\let\subparagraph\llncssubparagraph
\titlespacing*{\section}{0pt}{0.8\baselineskip}{0.8\baselineskip}
\titlespacing*{\subsection}{0pt}{0.5\baselineskip}{0.5\baselineskip}
\titlespacing*{\subsubsection}{0pt}{0.5\baselineskip}{0.5\baselineskip}

\begin{document}
\mainmatter              
\title{Contact Inertial Odometry: Collisions are your Friends}

\titlerunning{Contact Inertial Odometry: Collisions are your Friends}  
%
\author{Thomas Lew\inst{1}\thanks{Both authors contributed equally to this manuscript.} \and Tomoki Emmei\inst{2}\footnotemark[1] \and David D. Fan\inst{3} \and 
Tara Bartlett\inst{3} \and Angel Santamaria-Navarro\inst{3} \and Rohan Thakker\inst{3} \and Ali-akbar Agha-mohammadi\inst{3}}
%
\authorrunning{Thomas Lew et al.} 
%
\tocauthor{Thomas Lew*, Tomoki Emmei*, David D. Fan, Tara Bartlett, Rohan Thakker, Ali-akbar Agha-mohammadi}
%
\institute{ETH Z\"urich,  Switzerland\\
\email{thomas.lew@stanford.edu}
\and
The University of Tokyo, Japan
\and
NASA Jet Propulsion Laboratory, California Institute of Technology, USA
}

\maketitle              

\vspace{-0.25cm}
\begin{abstract}
\renewcommand{\baselinestretch}{1.0} 
Autonomous exploration of unknown environments with aerial vehicles remains a challenge, especially in perceptually degraded conditions. 
Dust, fog, or a lack of visual or LiDAR-based features results in severe difficulties for state estimation algorithms, which failure can be catastrophic.  
In this work, we show that it is indeed possible to navigate in such conditions without any exteroceptive sensing by exploiting collisions instead of treating them as constraints.
To this end, we present a novel contact-based inertial odometry (\textbf{CIO}) algorithm: it uses estimated external forces with the environment to detect collisions and generate pseudo-measurements of the robot velocity, enabling autonomous flight. 
To fully exploit this method, we first perform modeling of a hybrid ground and aerial vehicle which can withstand collisions at moderate speeds, for which we develop an external wrench estimation algorithm. 
Then, we present our CIO algorithm 
and develop a reactive planner and control law which encourage exploration by bouncing off obstacles. 
All components of this framework are validated in hardware experiments and we demonstrate that a quadrotor can traverse a cluttered environment using an IMU only. 
This work can be used on drones to recover from visual inertial odometry failure or on micro-drones that do not have the payload capacity to carry cameras, LiDARs or powerful computers.
\\

\textbf{Video:} Experimental results are available at \href{https://youtu.be/AGyu9tkhSLk}{https://youtu.be/AGyu9tkhSLk}
\renewcommand{\baselinestretch}{0.9} 
\end{abstract}

\vspace{-0.2cm}


	\begin{figure}[htb!]
		\centering
		\includegraphics[width=0.65\linewidth]{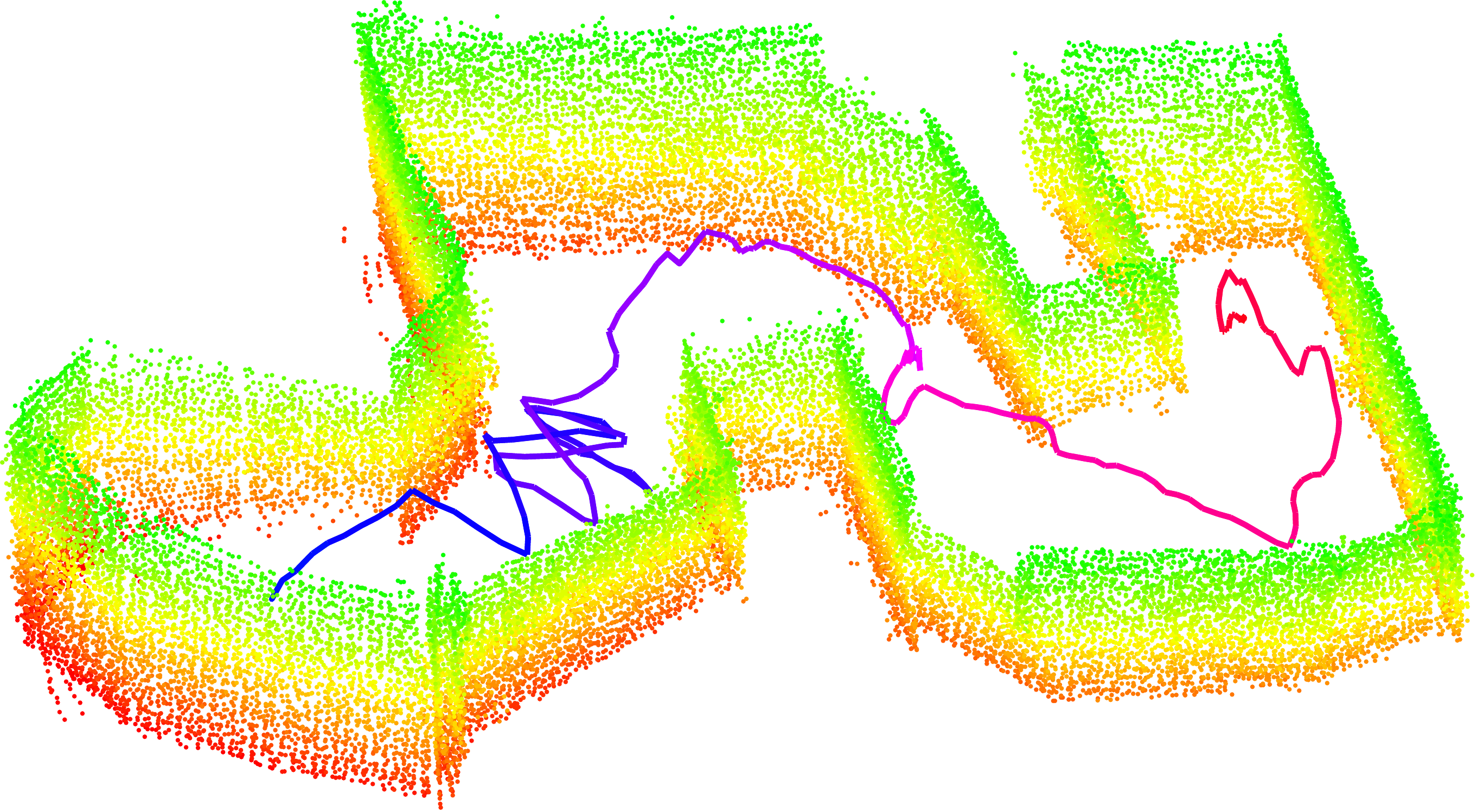}
		\caption{Blind aerial navigation in a cluttered environment.  The path taken by the quadrotor is indicated by the blue to pink line.  Multiple collisions with the environment (pointcloud) are leveraged to improve the velocity estimates of the vehicle and to traverse the maze. Using our contact inertial odometry (CIO) algorithm, the velocity of the robot can be estimated and controlled, enabling safe collisions for the vehicle.}
		\label{fig:blam_path}
		\vspace{-0.5cm}
	\end{figure}


\section{Introduction}
Collision avoidance has been a consistent theme in the robotics community since its inception.  In the motion and trajectory planning literature for instance, the goal often consists of computing a trajectory avoiding all obstacles which are deemed capable of harming the system.  This is especially true for applications such as aerial vehicles \cite{richter2016polynomial} 
and autonomous cars \cite{Paden2016ASO}.  However, contact with the environment can be highly informative, providing useful information for planning, control, and state estimation.  

Currently, autonomous navigation in \textit{perceptually degraded environments} is a challenge.
Dust, smoke, fog, and a lack of visual or LiDAR-based features result in severe difficulties for state estimation. As the errors in state estimates propagate to control and planning processes of the robot, such errors can be catastrophic for autonomous systems. 
%
%
%
%
There is a large body of work on accurate odometry techniques relying on \textit{exteroceptive} sensors (e.g. stereo cameras, LiDARs, radar, thermal cameras, GPS, etc.). However, such sensing modalities may fail in perceptually degraded conditions, e.g., when flying through dust.  
%
In contrast, \textit{proprioceptive} sensors (accelerometers, gyroscopes) are often much smaller, lighter, relatively cheaper and work regardless of assumptions on the environment. 
However, the full state of the robot is unobservable using an inertial measurement unit (IMU) only, which 
renders conventional IMU-only estimation methods insufficient.
Nevertheless, as reliable velocity estimates can be sufficient for navigation, we propose a novel velocity estimation, planning and control framework which exploits the information from contacts to enable IMU-based navigation.  
External force estimation methods have been extensively studied and can be generally divided between methods utilizing additional exteroceptive 
sensors 
\cite{Briod2013,Tomic2017}
and methods using proprioceptive sensors but relying on accurate state estimation and dynamics modeling \cite{Oh2017}.  
Using these methods, it is possible to estimate the position of contact points \cite{petrovskaya_global_2011}, 
enabling multiple applications.  
For instance, work in manipulation enabled 3d shape reconstruction of an object using repeated contacts \cite{inaba_manifold_2016}. 
When navigating in an environment with limited visibility and in cases where exteroceptive are either unavailable or fail, the methods above will be inaccurate, causing performance degradation or termination of the robot's operation.
%
%
For the above-mentioned reasons, utilizing contact information for state estimation can be beneficial.
	Contacts have been used in the legged robotics research within Kalman filters
\cite{kuindersma_optimization-based_2016}.
Using additional force sensors on each foot, assuming no slip for each foot in contact with the ground, and leveraging the forward kinematics of the system,
it is possible to provide feet position measurements to update the full state of the robot. 
Similarly, work in state estimation for smartphones performs zero velocity measurement updates when detecting that the user stops walking \cite{wagstaff_lstm-based_2018}.
However, such methods often assume the availability of a GPS system to detect the full stop of the user and perform the measurement update.
Pseudo measurements are also used in rolling systems, where the no-slip holonomic constraint can be leveraged \cite{Dissanayake2001} to update the velocity perpendicular to the driving direction. 

For aerial vehicles navigating in challenging environments, most of these assumptions do not hold: exteroceptive sensor may become unavailable or fail, collisions occur almost instantaneously and no passive force control to maintain the contact exists, as opposed to walking robots.
Recent work presented an approach to include contact information within a factor graph \cite{nisar_vimo_2019}, but assumes the availability of a camera and force sensor. 
Also, \cite{abeywardena_improved_2013,svacha2019inertial} developed IMU-only estimation methods for drones, but these approaches rely on estimating drag forces from rotor speeds which may be unobservable at low velocity. \\

\textbf{Contributions:} 
Instead of improving existing state estimation algorithms or achieving accurate state estimation assuming favorable flight conditions, the goal of this work is to leverage collisions to perform reasonable state estimation and enable autonomous robust navigation in challenging environments where exteroceptive sensors fail, e.g., when all visual sensors have failed due to dust, fog, smoke, or lack of features.  
We argue that by using resilient hardware which can withstand collisions at moderate speeds, colliding with obstacles becomes a valuable asset. 
To illustrate this claim, we present a novel measurement model to exploit dynamic contact information as a \textit{pseudo} velocity measurement which can be incorporated in an Extended Kalman Filter. We name the resulting odometry algorithm \textbf{CIO}: a novel contact inertial odometry algorithm which we couple with a reactive velocity planner to enable autonomous navigation in challenging environments. Compared to existing approaches, our approach only requires the estimation of the orientation of the contact force for state estimation and reactive planning. To the best of our knowledge, this is the first work which performs autonomous flight in a cluttered environment using proprioceptive sensors only.\vspace{-0.25cm}

\begin{figure}[!htb]
\begin{minipage}{.5\linewidth}
  \centering
		\includegraphics[clip, trim=2.4cm 0.0cm 1.1cm 0.2cm, 
		height=3.6cm
		]{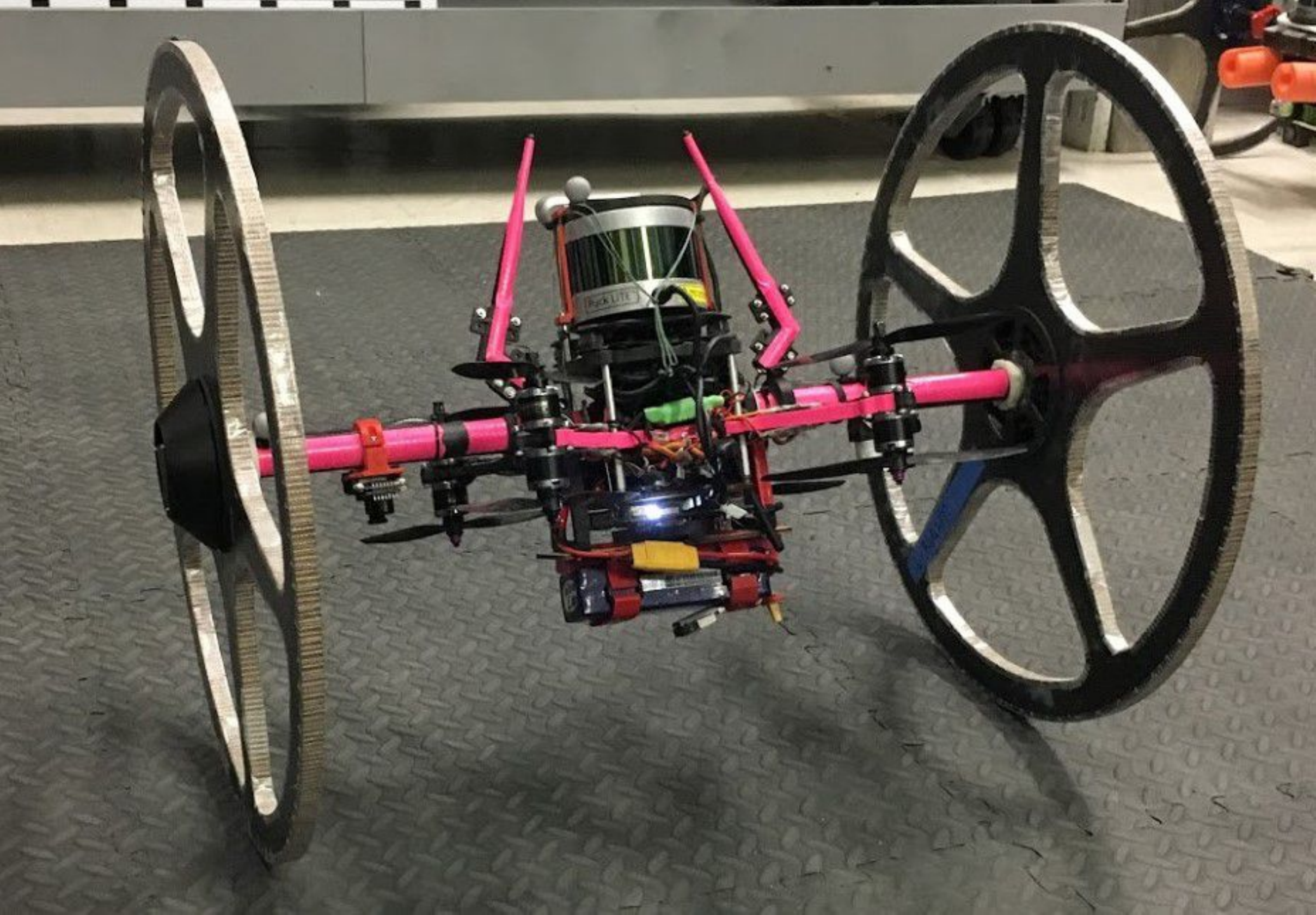}
\end{minipage}%
\begin{minipage}[t]{.03\linewidth}
  \centering  
  \includegraphics[width=1.0\linewidth]{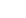}
\end{minipage}%
\begin{minipage}{0.47\linewidth}
  \centering
    \includegraphics[height=3.6cm
    ]{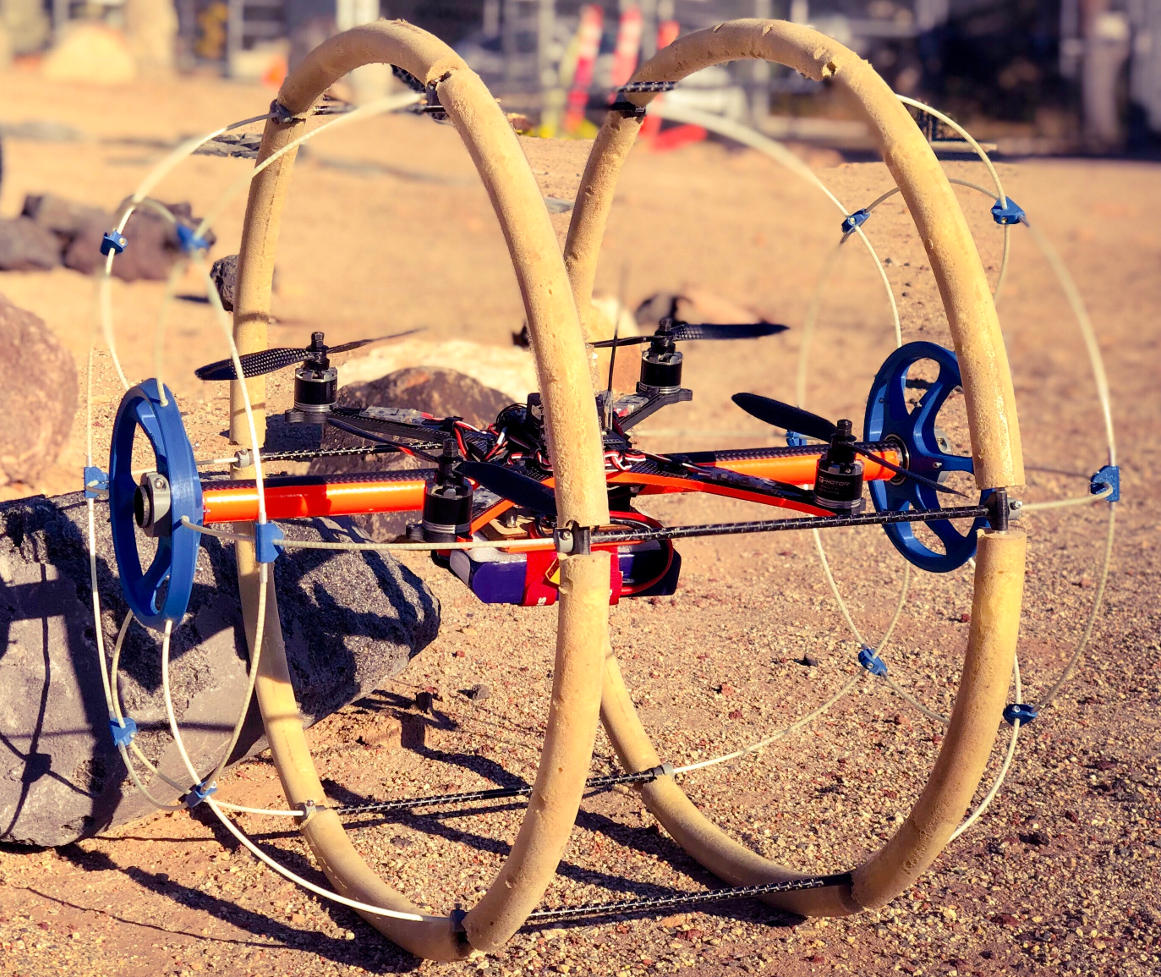}
\end{minipage}
\par
\vspace{0.5mm}
\begin{minipage}{1.0\linewidth}
  \centering
		\vspace{0.4cm}
		\caption{Rollocopter platform (left) and 
  Hytaq quadrotor (right). Our framework is applicable to any flying vehicle which can cope with collisions at moderate speeds.}
		\label{fig:polaris}
		\vspace{-0.75cm}
\end{minipage} 
\end{figure}

In order to fully demonstrate the capabilities of this contact-aware navigation, we leverage a hybrid ground and aerial vehicle: the \textit{Rollocopter} \cite{fanrollo,kalantari_spenko_2015,rollocopter_aeroconf2019}, shown in Figure \ref{fig:polaris}. By using two passive wheels attached to a quadrotor platform, this vehicle is capable of both rolling and flying, while being robust to collisions at moderate speeds. Such hybrid systems are well suited to our framework, since rolling can be thought of as an extended collision with the ground. 
%
For this class of vehicles, we provide new methods for contact force and point estimation.  
This enables our system to react to external obstacles and decide whether to roll or fly to traverse its environment.
Finally, CIO is leveraged to enable dynamic flying-bouncing behaviors, where the robot flies and periodically touches the ground. 
%
Although we perform analysis and experiments specific to the Rollocopter platform, we stress that in general, CIO is a powerful tool for any collision-resistant autonomous drone design equipped with an IMU, from lightweight quadrotors with propeller guards, to other types of hybrid vehicles \cite{drivocopter_aeroconf2020}.

The paper is organized as follows:  In Section \ref{sec:modeling_force_estimation} we leverage existing work in force estimation and extend it for our novel hybrid vehicle. By precisely describing the dynamics of our system, we achieve reliable collision detection, accurate force estimation and precise contact position estimation.  In Section \ref{sec:cio}, we present CIO: a novel IMU-only Contact Inertial Odometry  algorithm. We also propose a reactive planning and control strategy to traverse a cluttered environment.  Section \ref{sec:results} presents experimental validation of our approach.  Autonomous navigation in a dark and cluttered environment is demonstrated by flying, bouncing, and rolling.  Finally, we conclude in Section \ref{sec:conclusion} and discuss future directions of research for various research communities.

\section{System Modeling and Contact Forces Estimation}\label{sec:modeling_force_estimation}

\begin{table}[t]
	\centering
	\begin{tabular}{lcc}
		\hline 
		Variable & Parameter/Notation & Value \\ 
		\hline 
		\hline 
		Total and wheel masses & $m_t,m_{\rm w}$ & 4.036, 0.283 kg \\ 
		\hline 
		Total inertia & $I_t^x$, $I_t^y$, $I_t^z$ & 0.09, 0.074, 0.09 kg$\cdot$m$^2$ \\ 
		\hline 
		Body inertia & $I_b^x$, $I_b^y$, $I_b^z$ & 0.035, 0.0545, 0.035 kg$\cdot$m$^2$ \\ 
		\hline 
		Propeller diameter & $D$ & 0.2286 m \\ 
		\hline 
		Thrust and torque coefficients & $C_p,C_q$ & 0.11, 0.008 \\ 
		\hline 
		Air density & $\rho$ & 1.18 \\ 
		\hline 
		Wheel radius & $R$ & 0.2667 m \\ 
		\hline 
		Wheel inertia & $I_{\rm w}$ & 0.00975 kg$\cdot$m$^2$ \\ 
		\hline 
		Arm length and half length of the shaft &$l,L$ & 0.254, 0.3125 m \\
		\hline 
		Force, torque and wheel torque gains& $K_{F},K_{M},K_w$ & 10,10,10 \\ 
		\hline 
		Wheel side: left, right & $i \in \{l, r\}$ & \\
		\hline
		Rollocopter part: body, wheel, total & $p \in \{b, w, t\}$ & \\
		\hline
		Wrench: external, input and dragging & $k \in \{e, in, d\}$ & \\
		\hline
		Internal and external forces & $\mathbf{F}_{in},\mathbf{F}_{e} \in \mathbb{R}^3$ & \\
		\hline
		Internal and external moments & $\mathbf{M}_{in},\mathbf{M}_{e} \in \mathbb{R}^3$ & \\
		\hline
		External moments on left and right wheels& $M_{\rm w}^l,M_{\rm w}^r \in \mathbb{R}$ & \\
		\hline
		Linear and angular body velocities & $\mathbf{v},\boldsymbol\omega \in\mathbb{R}^3$ & \\
		\hline
		Wheel angular velocity on wheels & $\gamma^l,\gamma^r \in \mathbb{R}$ & \\
		\hline
		Contact force on wheels & $\mathbf{f}^l_e,\mathbf{f}^r_e\in\mathbb{R}^3$ &\\
		\hline
		Contact position on each wheel& $\mathbf{p}^i\in\mathbb{R}^3$ &\\
		\hline
		\vspace{0.2cm}
	\end{tabular}
	\caption{Parameters, variables, symbols and notations.}
	\label{tab:spec_rollo}
		\vspace{-0.5cm}
\end{table}


\subsection{System Modeling}

\subsubsection{Standard Quadrotor Platform}

We start with the modeling of the dynamics of a standard quadrotor platform, which we later modify for our hybrid vehicle. Omitting position and orientation and analyzing the motion of the vehicle in the body frame, a quadrotor vehicle of mass $m$ and inertia $\mathbf{I}$ can be described by its state $\mathbf{x}=[\mathbf{v};\boldsymbol\omega]\in\mathbb{R}^6$, where $\mathbf{v}$ and $\boldsymbol\omega$ are the linear and angular velocities, as
\begin{subequations}\label{eq:dynamics_quadrotor}
	\begin{align}
&m (\dot{\mathbf{v}}+\boldsymbol\omega \times \mathbf{v})=\mathbf{F}_{in}+\mathbf{F}_{e} \label{eq:mdl_body_trans}\\
&\mathbf{I}\dot{\boldsymbol\omega} + \boldsymbol\omega\times\mathbf{I}\cdot\boldsymbol\omega = \mathbf{M}_{in} + \mathbf{M}_{e}. \label{eq:model_rot}
    \end{align}
\end{subequations}

	
	The input wrench $\{\mathbf{F}_{in},\mathbf{M}_{in}\}$ can be computed as a function of the angular velocities $\bar{n}_j$ of the propellers. In the following derivations, we assume a standard quadrotor configuration with 8 propellers, as in our hybrid platform shown in Figure \ref{fig:thrustquad}. 
	The thrust and rotational torque of the $j$th proppeller can be described as $\rho C_pD^4\bar{n}_j^2 =	C_T\bar{n}_j^2$ and $\rho C_q D^5 \bar{n}_j^2 = C_Q\bar{n}_j^2$, respectively. 
%
%
%
The input force
	$\mathbf{F}_{in}=[0,0,F_{in}^z]^{T}$ and torque $\mathbf{M}_{in}=[M_{in}^x,M_{in}^y,M_{in}^z]^T$ can be expressed as
	\begin{equation}
	    \begin{bmatrix}
	    F_{in}^z \\ M_{in}^x \\ M_{in}^y \\ M_{in}^z
	    \end{bmatrix}
	    =
	    \begin{bmatrix}
			C_T & C_T & C_T & C_T & C_T & C_T & C_T & C_T \\ 
			-lC_T & lC_T & lC_T & -lC_T & lC_T & -lC_T & -lC_T & lC_T \\ 
			-lC_T & -lC_T & lC_T & lC_T & -lC_T & -lC_T & lC_T & lC_T \\ 
			-C_Q & C_Q & -C_Q & C_Q & C_Q & -C_Q & C_Q & -C_Q
	    \end{bmatrix}
	 \begin{pmatrix}
	\bar{n}_1^2\\
	\vdots\\
	\bar{n}_8^2
	\end{pmatrix} 
	.
		\end{equation}

\subsubsection{Multi-body Modeling of a Hybrid Vehicle} 
	\begin{figure}[htb!]
		\centering
		\includegraphics[width=60mm]{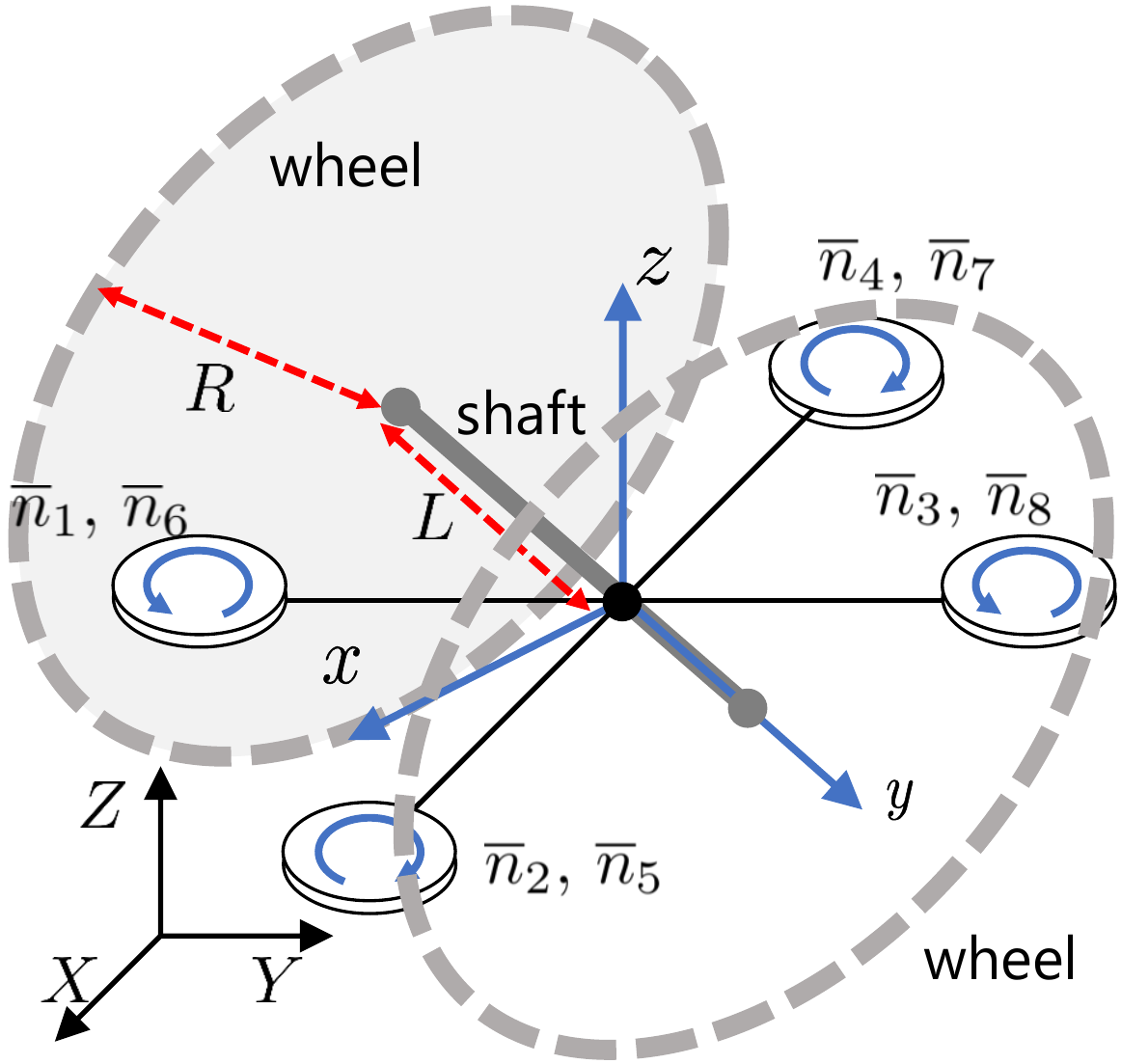}
		\caption{Model of the Rollocopter (with 8 propeller configuration).}
		\label{fig:thrustquad}
	\end{figure}

For the experiments presented in this paper, we leverage the Rollocopter: a hybrid rolling and flying vehicle shown in Figures \ref{fig:polaris} and \ref{fig:thrustquad}, with all variables and parameters specified in \tab{spec_rollo}.
	Using Kane's method \cite{kane_roithmayr_dynamics:_2016} and denoting the left and right wheel angular velocities as $\gamma_l$ and $\gamma_r$, the dynamics of the Rollocopter  can be expressed as
	\begin{subequations}\label{eq:dynamics_flying}
			\begin{align}
		&m_t(\mathbf{\dot{v}}+\boldsymbol\omega\times \mathbf{v})=\mathbf{F}_{in}+\mathbf{F}_{e} \label{eq:mdl_body_trans}\\
		&\mathbf{I}_t\dot{\boldsymbol\omega} + \boldsymbol\omega\times\mathbf{I}_b\cdot\boldsymbol\omega 
		+ \begin{pmatrix}
		2m_{\rm w}L^2(\dot{\omega}_x+\omega_z\omega_y) \\ 
		I_{\rm w}(\dot{\gamma}_l+\dot{\gamma}_r) \\ 
		2m_{\rm w}L^2(\dot{\omega}_z - \omega_x\omega_y)
		\end{pmatrix}
		= \mathbf{M}_{in} + \mathbf{M}_{e} \label{eq:model_rot} \\
		&I_{\rm w}(\dot{\omega}_y+\dot{\gamma}_i) = \hat{M}_{\rm w}^{i}, & i\in\{l,r\}, \label{eq:mdl_wheel}
		\end{align}
	\end{subequations}
    where $\mathbf{I}_t, \mathbf{I}_b$ and $\mathbf{I}_{\rm w}$ denote the total, body and wheel inertias, and $m_t,m_{\rm w}$ the total and wheel masses, respectively. Using these equations, it is possible to estimate the external wrench $\{\mathbf{F}_{e},\mathbf{M}_{e}\}$ acting on the vehicle and the positions of contacts on the wheels.

	\subsection{External Wrench Estimation}\label{sec:external_force_est}
	In this section, the dynamical model of the hybrid vehicle 
    is used to estimate the external contact wrench in both flying and rolling modes.
    This is achieved with proprioceptive sensors only,  using accelerometer measurements $\mathbf{a}= \mathbf{\dot{v}}+\boldsymbol\omega\times \mathbf{v}$, gyroscope measurements $\boldsymbol\omega$, and wheel angular velocity and acceleration measurements  $(\gamma_r,\gamma_l)$ and $(\dot\gamma_r,\dot\gamma_l)$.  
    For a standard quadrotor platform, or in the absence of wheel encoders, similar estimation equations can be derived by setting $m_{\rm w}\approx 0$ and $I_{\rm w}\approx 0$. In this work, it is assumed that no other external disturbances (e.g. wind) act on the system. Also, we avoid the use of a high pass filter to enable the estimation of contact forces in situations where the system starts in contact with its environment.
	
	\subsubsection{Flying Mode}
	
	
	Based on the dynamical equations of the system in \eqref{eq:dynamics_flying} and following a similar approach as in \cite{Tomic2017}, we derive a residual vector of the dynamics and use a first-order low-pass filter to estimate the external wrench $\{\hat{\mathbf{F}}_e, \hat{\mathbf{M}}_e, \hat{M}_{\rm w}^l, \hat{M}_{\rm w}^r\}$ as
	\begin{subequations}
		\begin{align}
			\hat{\mathbf{F}}_e &= K_{F}\int_0^t(m {\cdot}\mathbf{a} -\mathbf{F}_{in}-\hat{\mathbf{F}}_e) \ \mathrm{dt}
			\label{eq:f_es_imu}
			\\
			\mathbf{\hat{M}_e} &= K_{M}\left(\mathbf{I}_t\boldsymbol\omega+
			\displaystyle\bigintsss_0^t \left(
			\boldsymbol\omega\times\mathbf{I}_b \cdot \boldsymbol\omega +
			\begin{pmatrix}
				2m_{\rm w}L^2(\dot{\omega}_x{+}\omega_z\omega_y) \\ 
				I_{\rm w}(\dot{\gamma}_l{+}\dot{\gamma}_r) \\ 
				2m_{\rm w}L^2(\dot{\omega}_z {-} \omega_x\omega_y)
			\end{pmatrix}
			- \mathbf{M}_{in} - \mathbf{\hat{M}_{e}}
			\right)\mathrm{dt}
			\right)
			\\
			\hat{M}_{\rm w}^{i} &= K_{\rm w}\left(I_{\rm w}(\omega_y+\gamma_i)-\int^t_0\hat{M}_{\rm w}^{i} \ \mathrm{dt}
			\right), \qquad\qquad\qquad\qquad\qquad\qquad i\in\{l,r\}
			.
			\label{eq:tau_wei_integrated}
		\end{align}
	\end{subequations}

	\subsubsection{Rolling Mode}
	Using the wheels of the hybrid vehicle to drive on the terrain, and assuming no wheels slip, the following non-holonomic constraints can be derived:
	\begin{align}
		v_x = \frac{R}{2}(\gamma_r + \gamma_l+2\omega_y),\ \ 
		v_y = 0,\ \ 
		\omega_z = \frac{R}{2L}(\gamma_r - \gamma_l)
		.
		\label{eq:NonHolo}
	\end{align}
	The derivation of (\ref{eq:NonHolo}) can be found in the Appendix \ref{apdx:sec:nonholonomic_constraints}\footnote{The appendix is available at \href{https://arxiv.org/abs/1909.00079}{https://arxiv.org/abs/1909.00079}}.
    Furthermore, since the reference frame of the robot is defined with respect to the ground, $v_z=0$ and $\omega_x=0$ hold. Using these additional constraints, the residual error  from the constrained dynamics (see Appendix \ref{apdx:sec:force_est_rolling_mode}) can be used 
    to derive the following external wrench estimation equations: 
\begin{subequations}
	\begin{align}
		\hat{F}_{e}^x &= K_{F}\displaystyle\bigintssss_0^t\left(\left(\frac{m_tR}{2} + \frac{2I_{\rm w}}{R}\right)(\dot\gamma_r + \dot\gamma_l + 2\dot{\omega}_y) - F_{in}^x - \hat{F}_{e}^x\right) \ \mathrm{dt}
		\\
		\hat{F}_{e}^y &= K_{F}\displaystyle\bigintssss_0^t\left(\frac{m_tR^2}{4L}(\gamma_r - \gamma_l)(\gamma_r + \gamma_l +2\omega_y) - \hat{F}_{e}^y\right) \ \mathrm{dt}\\
		\hat{M}_{e}^z &= K_{M}\displaystyle\bigintssss_0^t\left(\left(\frac{R}{2L}I_t^z+m_{\rm w}LR+\frac{I_{\rm w}L}{R}\right)(\dot\gamma_r - \dot\gamma_l) - M_{in}^z - \hat{M}_{e}^z\right) \ \mathrm{dt}
		.
	\end{align}
\end{subequations}

	
	\subsection{Contact Point Estimation}
	The position of the contact on the wheel while rolling is important to decide whether to fly or to roll. 
	%
	Estimating this position can be written as an optimization problem which can be solved analytically, with the known total external wrench $\mathbf{W}_e\in\mathbb{R}^8$ and unknown variables $\boldsymbol\zeta\in\mathbb{R}^{12}$ given as\\[-1mm]
	\begin{equation*}
	    \mathbf{W}_e = [\mathbf{F}_e, \mathbf{M}_e, M_{\rm w}^l, M_{\rm w}^r]
	    ,
	    \quad 
	    \boldsymbol\zeta = [\mathbf{f}_e^l,\mathbf{f}_e^r, \mathbf{p}^l, \mathbf{p}^r],
	\end{equation*}\\[-1mm]
	where $\mathbf{f}_{e}^i=[{f}_x^i,{f}_y^i,{f}_z^i]$ and $\mathbf{p}^i=[p_x^i,p_y^i,p_z^i]$, $i\in\{l,r\}$ denote the contact external forces and contact points on the left and right wheels. 
%
%
%
%
%
%
%
%
	To solve for $\boldsymbol\zeta$, we make the following assumptions and obtain the corresponding constraints:\\[-14pt]
%
%
%
%
%
	\begin{enumerate}
	    \item Contacts only occur on wheels and the terrain is flat: $p_y^i {=} 0$, $(p_x^i)^2 {+} (p_z^i)^2 {=} R^2, i\in\{ l, r\}$ and $\mathbf{F}_e = \mathbf{f}_e^l + \mathbf{f}_e^r + m_t \mathbf{g}$,
	    \item Collisions only occur on the left side ($f_y^r=0$) or on the right side depending on the sign of $F_e^y$,
	    \item Body and wheel external torques occurring on the wheels can be expressed as\vspace{-0.2cm}
	\end{enumerate}
    \noindent\begin{minipage}{.55\textwidth}
	\centering
	{\normalsize
	    \begin{align}
    		\mathbf{M}_{e}&=\begin{pmatrix}
    		{\displaystyle L(f_{\rm z}^l-f_{\rm z}^r)-p_{\rm z}^lf_{\rm y}^l - p_{\rm z}^rf_{\rm y}^r }\\ 
    		{\displaystyle M_{\rm w}^l + M_{\rm w}^r }\\ 
    		{\displaystyle L(f_{\rm x}^r-f_{\rm x}^l)+p_{\rm x}^lf_{\rm y}^l + p_{\rm x}^rf_{\rm y}^r}
    		\end{pmatrix} 
		\label{eq:model_rot_detailed}
	    \end{align} 
	}%
    \end{minipage}%
    \begin{minipage}{.10\textwidth}
    \centering
    \text{ }
    \end{minipage}%
    \begin{minipage}{.35\textwidth}
	\begin{subequations}\label{eq:torques_wheels_external}
	\centering
	{\normalsize
		\begin{align}
    		M_{\rm w}^l &= p_{\rm z}^lf_{\rm x}^l-p_{\rm x}^lf_{\rm z}^l
    		\label{eq:model_wheel_detailed}\\
    		M_{\rm w}^r &= p_{\rm z}^rf_{\rm x}^r-p_{\rm x}^rf_{\rm z}^r
    		.
    		\label{eq:model_wheel_detailed2}
		\end{align}
	}%
	\end{subequations}
    \end{minipage}
    \\[6pt]

    \eqref{eq:model_rot_detailed} can be derived using Kane's Equations. 
%
%
%
%
%
%
%
%
	Using these 12 constraints and by simultaneously solving these equations, we can compute a solution for $\boldsymbol\zeta$. The final expression and all derivations can be found in Appendix \ref{apdx:contact_point_position}.  

	\section{Contact-Based Odometry,  Planning and Control}\label{sec:cio}
	In the previous section, we presented a method to estimate the external wrench using proprioceptive sensors only. 
	As shown in this work, this information can be used to detect collisions and plan trajectories to bounce off walls. 
	However, to maintain stability of the drone, reliable velocity estimates are necessary.
	Unfortunately, velocity estimates obtained by propagating an IMU alone have unbounded drift, which could lead to catastrophic crashes at high speed.
	Therefore, to enable IMU-only navigation, we propose a novel Contact Inertial Odometry (CIO) algorithm which exploits contacts to reduce the error in velocity estimates,
	as well as a control and reactive planning strategy, enabling autonomous navigation in a cluttered environment without exteroceptive sensors.\\[-6mm]

	\begin{figure}
		\centering
		\includegraphics[width=\linewidth]{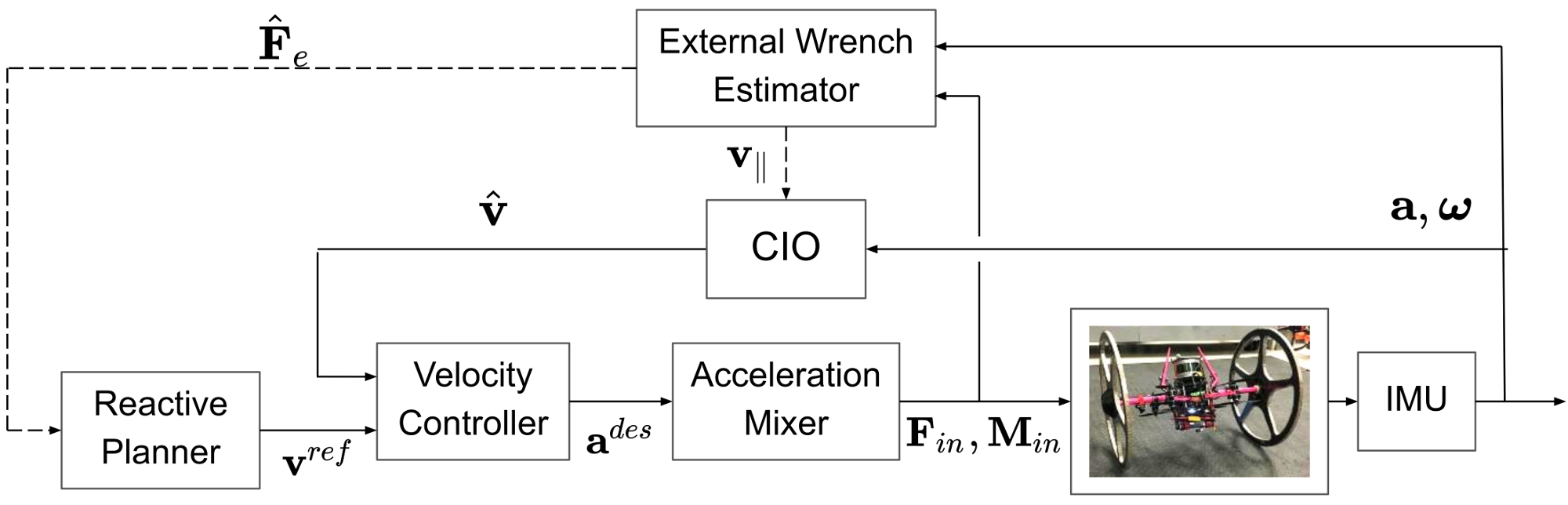}
		\vspace{-3mm}
		\caption{Diagram of the guidance, navigation and control architecture.}
		\label{fig:block}
	\end{figure}

	\subsection{Contact Inertial Odometry}\label{sec:cio_estimation}
	
	In this section, we present a simple method to include contact information as a measurement update within a Kalman Filtering framework. Given a robot pose described by its state $\mathbf{x}_k=[\mathbf{r}_k;\mathbf{q}_k;\mathbf{v}_k;\boldsymbol\omega_k]\in\mathbb{R}^{13}$ at time $k$, with $\mathbf{r}_k\in\mathbb{R}^3$ the robot position, $\mathbf{q}_k\in\mathbb{R}^4$ the quaternion describing its orientation and $\mathbf{v}_k,\boldsymbol\omega_k\in\mathbb{R}^3$ the linear and angular velocities respectively, the discrete time nonlinear dynamics of the system are assumed to be corrupted by Gaussian-distributed noise as
	$
	    \mathbf{x}_{k+1} = f(\mathbf{x}_k) + \boldsymbol\nu_k
	 $, where $\boldsymbol\nu_k\sim\mathcal{N}\left(\mathbf{0},\mathbf{Q}_k\right)$, with $\mathbf{Q}_k\succ 0$ the process noise covariance.  We then implement a standard Extended Kalman Filter (EKF), the prediction step of which is described in \cite{MooreStouchKeneralizedEkf2014}.
	Note that this EKF does not estimate biases of sensors, e.g., of the IMU.
	Given a measurement $\mathbf{z}_k$ of the state $\mathbf{x}_k$, corrupted by i.i.d. Gaussian-distributed noise $\mathbf{w}_k\sim\mathcal{N}\left(\mathbf{0},\mathbf{R}_k\right)$, with $\mathbf{R}_k\succ 0$, written as
	\begin{equation}
	    \mathbf{z}_k=h(\mathbf{x}_k) + \mathbf{w}_k
	    ,
	\end{equation}
	it is possible to perform a standard measurement update to the EKF as
	\begin{subequations}\label{eq:ekf_robot_local_update}
	\begin{align}
	\mathbf{K}_k &= \hat{\mathbf{P}}_k\mathbf{H}_k^T\left(\mathbf{H}_k\hat{\mathbf{P}}_k \mathbf{H}_k^T + \mathbf{R}_k \right)^{-1}
	\\
	\mathbf{x}_{k+1} &= \hat{\mathbf{x}}_{k+1} + \mathbf{K}_k(\mathbf{z}_k-\mathbf{H}_k\hat{\mathbf{x}}_k)
	\\
	\mathbf{P}_k &= (\mathbf{I} - \mathbf{K}_k\mathbf{H}_k) \hat{\mathbf{P}}_k (\mathbf{I} - \mathbf{K}_k\mathbf{H}_k)^T + \mathbf{K}_k\mathbf{R}_k\mathbf{K}_k^T 
	,
	\end{align}
	\end{subequations}
	where $\mathbf{H}_k$ denotes the Jacobian matrix of $h(\cdot)$, $\hat{\mathbf{P}}_k$ the predicted covariance of the predicted state $\hat{\mathbf{x}}_{k+1}$ and $\mathbf{I}$ the identity matrix.
	%
    The measurement model $h(\cdot )$ is used to encapsulate the information from a collision. During a contact, we assume the velocity of the robot to be parallel to the collided obstacle, with a velocity component perpendicular to the obstacle set to zero. Given a previously estimated velocity $\hat{\mathbf{v}}_{prev}=\hat{\mathbf{v}}_k$ and an estimate of the external force $\hat{\mathbf{F}}_e$, the parallel velocity $\mathbf{v}_{\parallelsum}$ is computed as
	\begin{align}\label{eq:vel_parallel}
		\mathbf{v}_{\parallelsum} = \hat{\mathbf{v}}_{prev} - \frac{(\hat{\mathbf{v}}_{prev}\cdot\hat{\mathbf{F}}_e)}{\hat{\mathbf{F}}_e\cdot\hat{\mathbf{F}}_e}\hat{\mathbf{F}}_e
		.
	\end{align}
	
	Given this parallel velocity at time $k$, we introduce a pseudo-measurement for the velocity of the system as
	\begin{equation}\label{eq:cio_meas_zero_vel}
	    \mathbf{z}_k = h(\mathbf{v}_k) + \mathbf{w}_k = \mathbf{v}_k + \mathbf{w}_k \gets \mathbf{v}_{\parallelsum}
	    ,
	\end{equation}
    where $\mathbf{v}_{\parallelsum}$ is computed according to \eqref{eq:vel_parallel} using the force estimate computed using \eqref{eq:f_es_imu}. 
	This measurement update is inspired from the literature in state estimation for driving vehicles \cite{Dissanayake2001}, where non-holonomic constraints enable pseudo-measurements in the driving direction. Similarly, in this work, we assume that the velocity perpendicular to the obstacle is zero, whereas the parallel components remains unaffected. This is based on the assumption that no energy is lost in the direction parallel to the obstacle, whereas the velocity is instantly zero at the time of the impact. 
	Including loss of energy due to friction in the parallel direction of the contact would require known properties of the wall, which are not necessarily available when operating in unknown environments. Since this loss of energy is proportional to the integral of the collision force, it would also require high accuracy force measurements at high rates, which are not necessarily available using proprioceptive sensors such as a low cost IMU. 

\begin{figure}[!htb]
\begin{minipage}{.49\textwidth}
  \centering
    \includegraphics[width=0.7\linewidth]{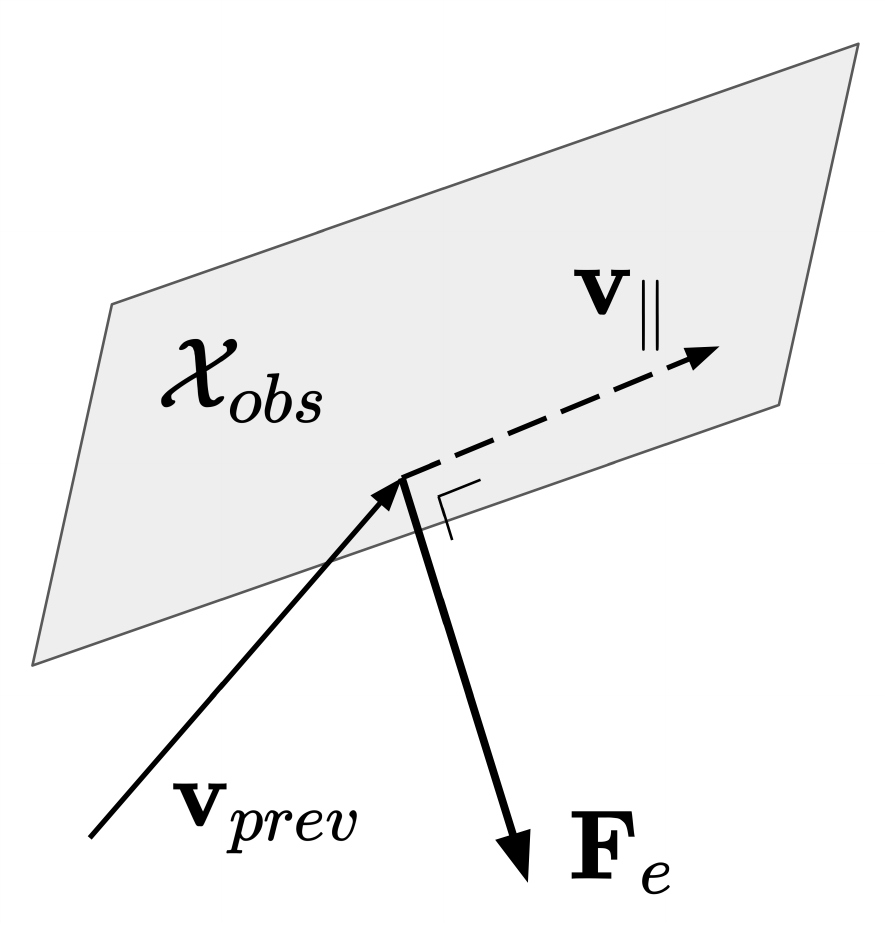}
\end{minipage}%
\begin{minipage}[t]{.02\textwidth}
  \centering  
\end{minipage}%
\begin{minipage}{0.49\textwidth}
  \centering
    \includegraphics[width=0.9\linewidth]{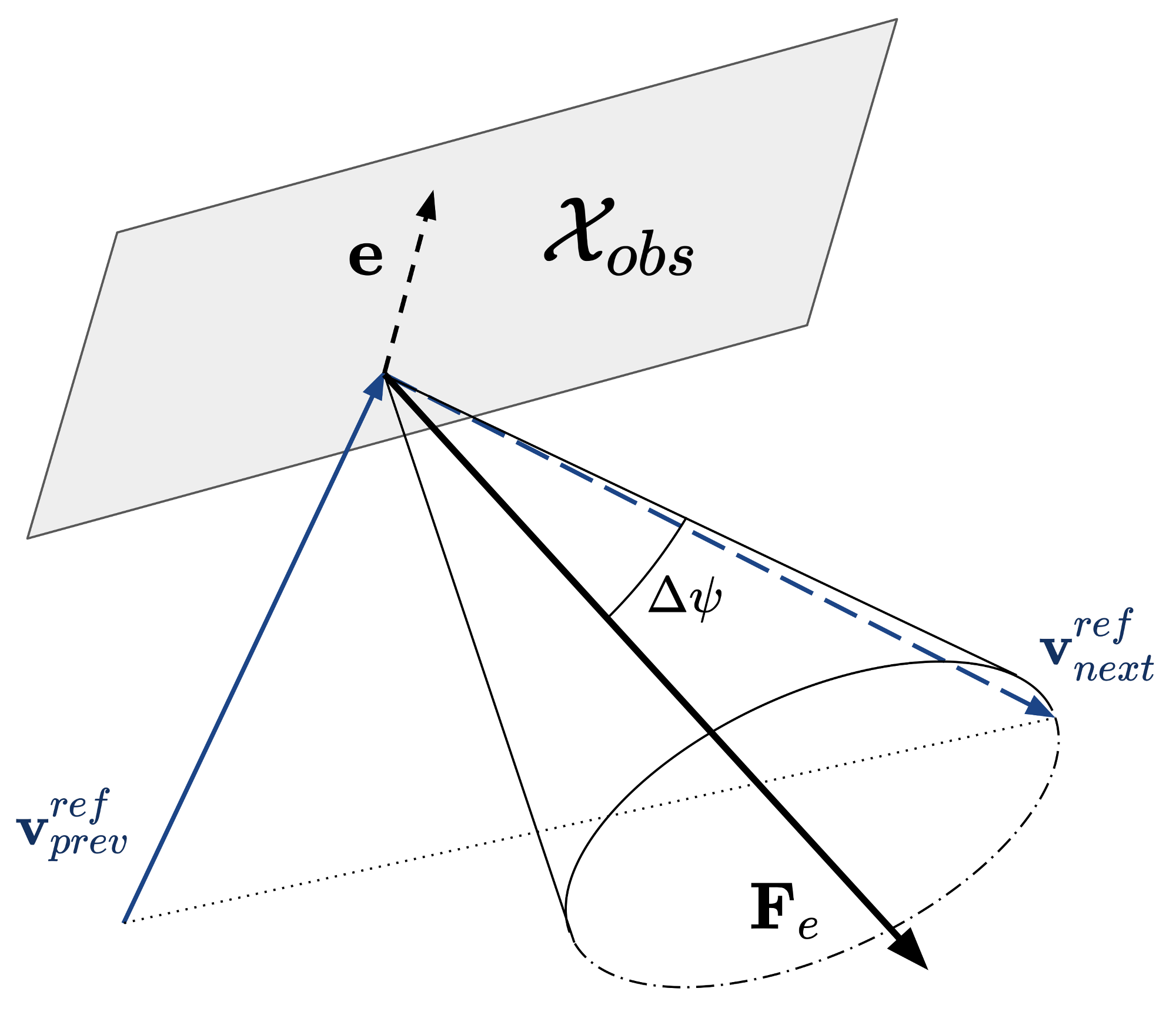}
\end{minipage}
\par
\begin{minipage}{0.48\textwidth}
  \centering
  \caption{Parallel velocity used as a pseudo-measurement after a contact.}
  \label{fig:par_vel_measurement}
\end{minipage}%
\begin{minipage}[t]{.04\textwidth}
  \centering  
  \includegraphics[width=1.0\linewidth]{fig/all/white.png}
\end{minipage}%
\begin{minipage}{0.48\textwidth}
  \centering  
  \caption{New desired velocity computed by our reactive planner after a contact.}
  \label{fig:velocity_ref}
\end{minipage} 
\vspace{-0.5cm}
\end{figure}
%
%


	\subsection{Reactive Planning}\label{sec:reactive_planner}
	
	Leveraging the proposed force estimation method, we present a reactive planner capable of generating reference velocities to navigate in unknown environments and react to collisions.
    %
    %
	Inspired by previous work on random sampling \cite{Gryazina2014}, a possible planning method could be the following:\\[-0.5cm]
	\begin{align}
		\mathbf{v}_{ref} = \hat{\mathbf{v}}_{prev} - 2\frac{(\hat{\mathbf{v}}_{prev}\cdot\hat{\mathbf{F}}_e)}{\hat{\mathbf{F}}_e\cdot\hat{\mathbf{F}}_e}\hat{\mathbf{F}}_e
		.
		\label{eq:bouncing}
	\end{align}

    It is possible to show that such a method is guaranteed to uniformly traverse an environment \cite{Gryazina2014}.
	However, it relies on accurate knowledge of the previous velocity $\hat{\mathbf{v}}_{prev}$, which may be inaccurate. Furthermore, for frontal collisions, the resulting direction would cause the system to return to its original position, which may not be adequate for exploration.  On the other hand, a wall-following strategy \cite{8793794} could be used, although its success would depend on the structure of the environment.

	We combine the advantages of these two approaches and propose a reactive planner based on the previously computed reference velocity $\mathbf{v}_{prev}^{ref}$ and estimated contact force $\mathbf{F}_e$. 
	As shown in Figure \ref{fig:velocity_ref}, the method consists of projecting $\mathbf{v}_{prev}^{ref}$ onto a cone around $\mathbf{F}_e$, defined with an angle $\Delta\psi$.  To include random sampling of the reference headings, we sample $\Delta\psi \sim \textrm{Unif}(\Delta\psi_{min},\Delta\psi_{max}) $, where $\textrm{Unif}(a,b)$ denotes the uniform distribution with values in $[a,b]$.
	To compute a reference direction $\tilde{\mathbf{v}}_{next}^{ref}$, we first compute a rotation axis $\mathbf{e}$ perpendicular to $\mathbf{F}_e$ and $\mathbf{v}_{prev}^{ref}$.
	Then, we use Rodrigues' rotation formula to rotate $\mathbf{F}_e$ by the angle $\Delta\psi$ around $\mathbf{e}$. Finally, we normalize the reference velocity and set its norm to a nominal velocity magnitude $v_{nom}$. These steps can be written as
	\begin{subequations}\label{eq:cone_planning}
	\begin{align}
	    \mathbf{e} &= \frac{\hat{\mathbf{F}}_e \times \mathbf{v}_{prev}^{ref}}{\|\hat{\mathbf{F}} \times \mathbf{v}_{prev}^{ref}\|}
	    \\
	    \tilde{\mathbf{v}}_{next}^{ref} &= \cos(\Delta\psi)\hat{\mathbf{F}}_e  + 
	    \sin(\Delta\psi)(\mathbf{e}\times\hat{\mathbf{F}}_e ) + 
	      ( 1{-}\cos(\Delta\psi) ) ( \mathbf{e} \cdot \hat{\mathbf{F}}_e )\ \mathbf{e}
	    \\
	    \mathbf{v}_{next}^{ref} &=  \frac{v_{nom}}{\|\tilde{\mathbf{v}}_{next}^{ref}\|} \tilde{\mathbf{v}}_{next}^{ref}
	      .
	\end{align}
	\end{subequations}

	Drift in the vertical velocity is particularly undesirable, but can be avoided by periodically making physical contact with the ground. 
	This behavior is implemented in addition to Equations \eqref{eq:cone_planning} and further demonstrates the capabilities of hybrid vehicles. 
	


	
	

	\subsection{Low-Level Controller}
	\fig{block} shows the block diagram of the control architecture. The controller receives a desired velocity $\mathbf{v}^{ref}$, which is generated by the reactive planner. 
	It is mapped to a desired acceleration with a proportional controller using the current estimated velocity.  Then, the desired acceleration and yaw (set to $0$) are mapped to a desired thrust and attitude quaternion via a geometric control method on SE(3) \cite{lee2010geometric}.  Finally, we rely on the on-board flight controller's attitude controller for tracking of the desired thrust and attitude.  This attitude controller runs at 200Hz 
	and makes use of the flight controller's attitude estimator, which generally produces reliable attitude estimates  since it is decoupled from the estimates of position and velocity.  It converts the desired thrust and attitude quaternion to four command inputs $F_{in,z},M_{in}^x,M_{in}^y,M_{in}^z$, which are then mapped to motor PWM commands.  This cascaded architecture works well because attitude and angular rate estimates are updated at a high frequency (200Hz) and are independent of position and velocity estimates. 
	

	\section{Experimental Results}\label{sec:results}
	
     We present experimental validation for the proposed force estimation and collision detection method, contact inertial odometry algorithm, and the reactive control and planning framework coupled with CIO.  All results were performed on the Rollocopter platform shown in Figure \ref{fig:polaris}. It is equipped with 
an Intel NUC i7 Core computer for on-board computation,
	an Intel RealSense RGBD camera, 
	a Garmin LiDAR-Lite range sensor, 
	a Pixhawk v2.1 flight controller with an on-board IMU which includes an accelerometer and a gyroscope, and
	hall effect wheel encoders.
%
%
%
	To show the applicability of our method on conventional quadrotors, we do not use wheel encoders and use the dynamics in \eqref{eq:dynamics_quadrotor} for the autonomous navigation experiments in Sections \ref{sec:results:cio_est} and \ref{sec:results:maze}.

	\subsubsection{Collision Detection}
	To detect collisions from estimated forces and trigger measurement updates and new reference velocities, we implement a thresholded detection.  To exploit all proprioceptive sensors on the Rollocopter, we introduce the following evaluation function $W[k]$:
		\begin{align}\label{eq:collision_threshold}
			W[k]=	w_{F_e}\|\hat{\mathbf{F}}_e[k]\|^2 + w_{M_e}\|\mathbf{\hat{M}_e}[k]\|^2 + w_{M_{\rm w}}(\hat{M}_{\rm w}^l[k]^2+\hat{M}_{\rm w}^r[k]^2) ,
		\end{align}
	where the hyperparameters are chosen as $w_{F_e} = 1$, $w_{M_{e}} = 1/L^2$, $w_{M_{\rm w}}\ = 1/R^2$ to ensure that $W[k]$ has the dimension of a force and simplify the tuning process.  A collision is detected when this value exceeds a threshold. 
	For the autonomous navigation experiments in Sections \ref{sec:results:cio_est} and \ref{sec:results:maze}, we set  $w_{M_{e}},w_{M_{\rm w}}=0$, as large contact forces dominate. 
    However, for estimating contact points on the wheels, the estimation of the torques on the wheels and body becomes more valuable.  Furthermore, as discussed in \cite{Briod2013}, larger distances between the contact point and the IMU deteriorates the quality of the estimation, which may require additional sensors (e.g. for Rollocopter, wheel encoders).

	\subsection{External wrench estimation for Rollocopter}\label{subsec:estimation_accuracy_improvement_based_on_multibody}


{
\begin{figure}[!htb]
\begin{minipage}{1.0\textwidth}
		\centering
		\vspace{-0.5cm}
		\subfigure{
			\includegraphics[width=38mm]{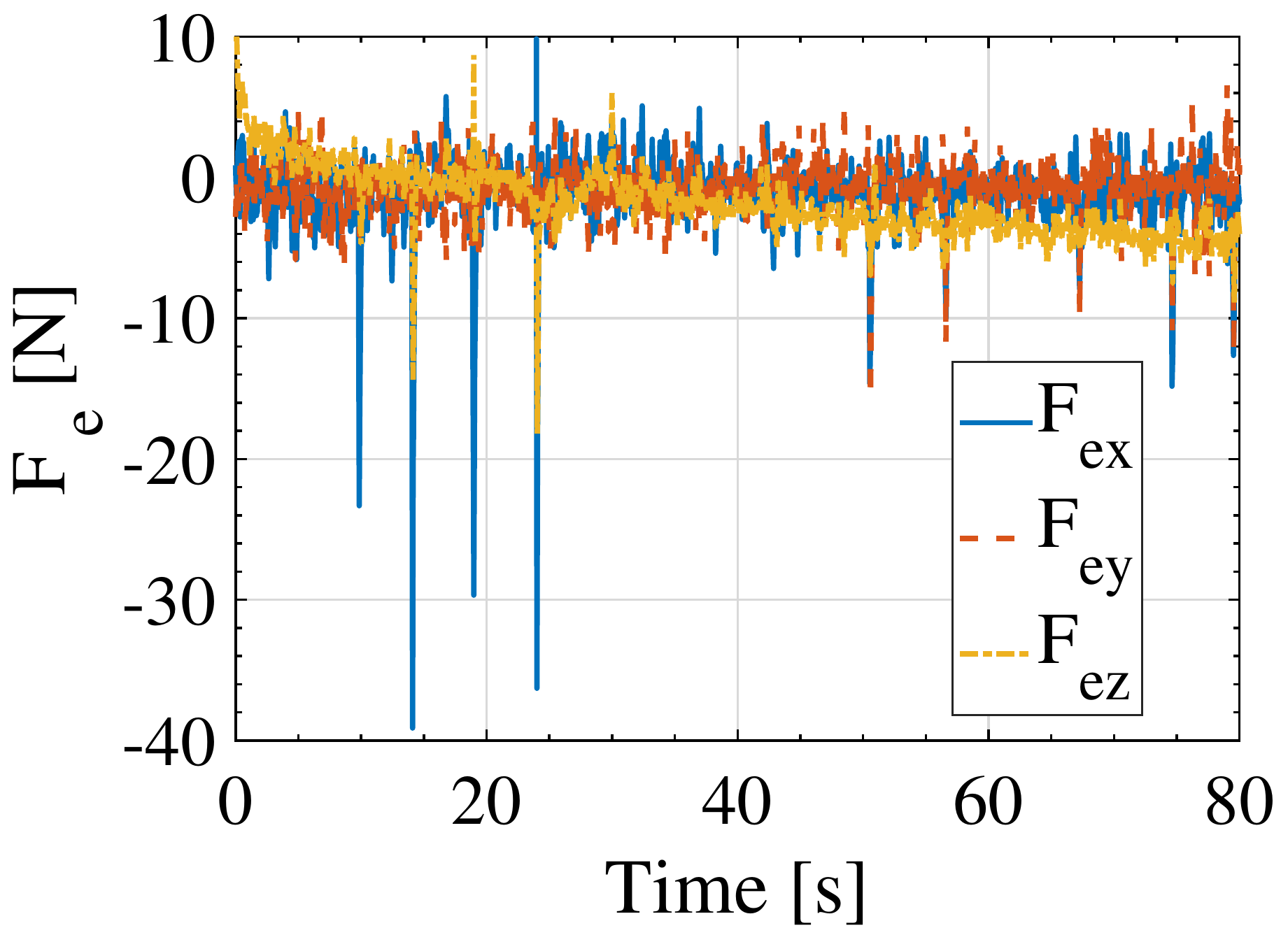}
			\label{fig:est_acc_Fe}}
		\subfigure{
			\includegraphics[width=38mm]{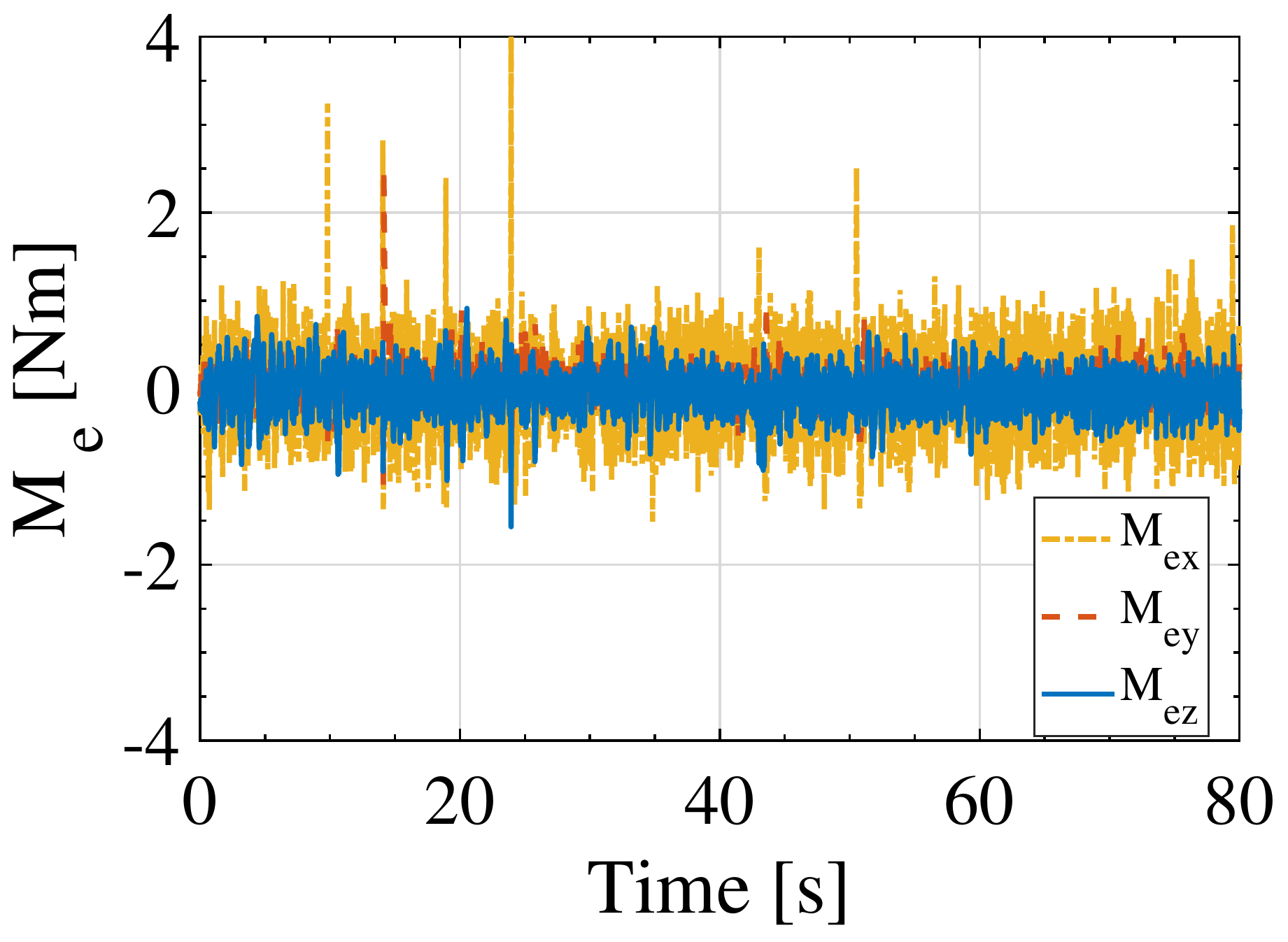}
			\label{fig:est_acc_Me}}
		\subfigure{
			\includegraphics[width=38mm]{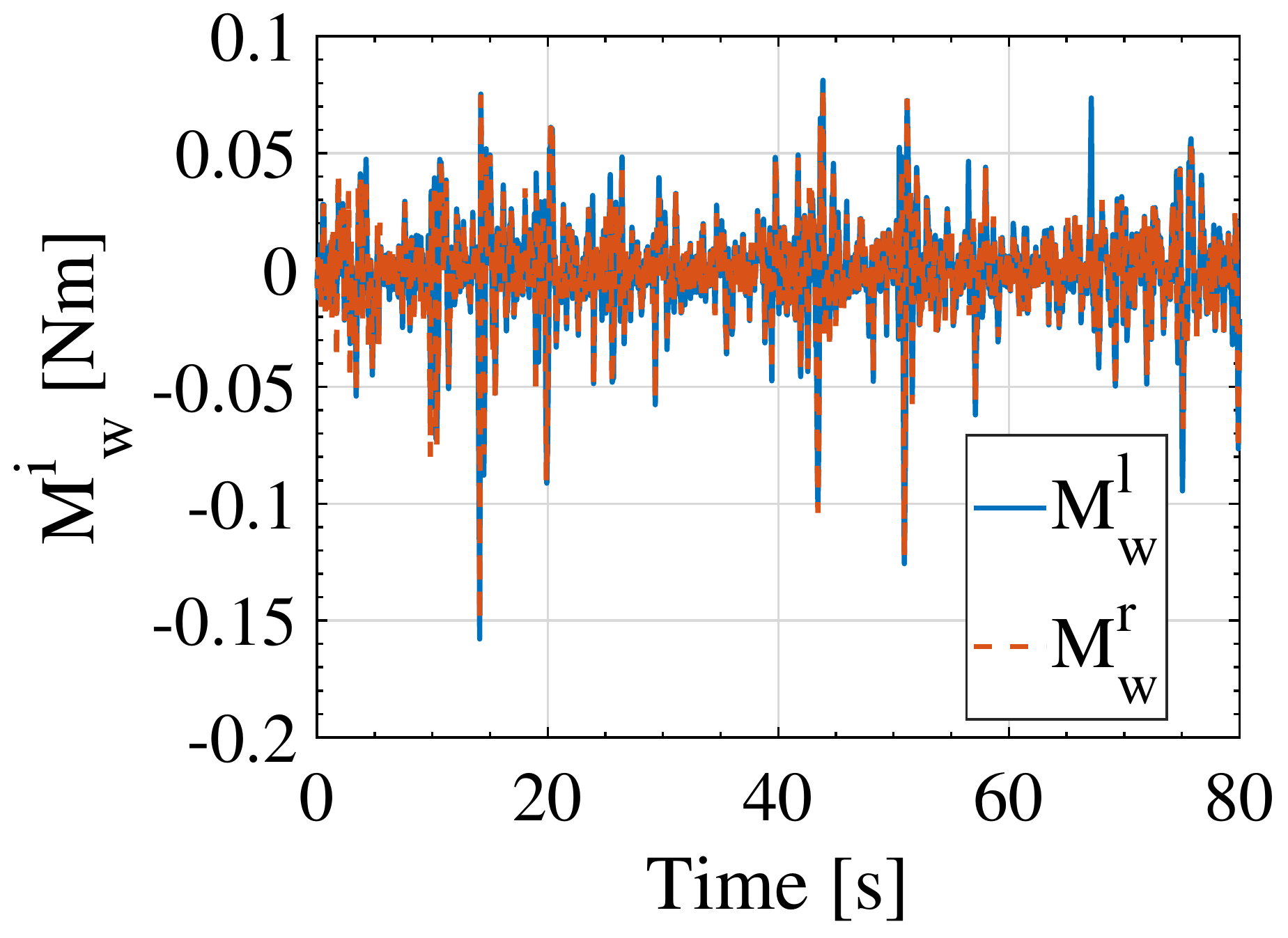}
			\label{fig:est_acc_Mwe}}
\end{minipage}%
\par
\begin{minipage}{1.0\linewidth}
  \centering
  \vspace{0.25cm}
		\begin{tabular}{|c||c|c|c|c|c|c|c|c|c|c|}
			\hline 
			time & 10s & 14s & 20s & 24s & 43s & 51s & 57s & 68s & 75s & 80s \\
			\hline
			direction & F(0$^\circ$)& F(0$^\circ$)& FD(0$^\circ$)& FU(0$^\circ$)& FL(45$^\circ$) & FL(45$^\circ$) & FL(45$^\circ$) & FL(45$^\circ$) & FL(45$^\circ$) & FL(45$^\circ$) \\
			\hline 
			estimated &$9.0^\circ$& $3.3^\circ$& $6.3^\circ$& $3.0^\circ$& - & $45.7^\circ$ & $49.43^\circ$ & $49.7^\circ$ & $32.4^\circ$ & $34.8^\circ$ \\
			\hline 
		\end{tabular} 
		\caption{Force estimation (top) and ground truth of collision times and directions (bottom) for the flying mode. Contact directions are denoted as F, B, D, U and L for frontal, backward, downward, upward and left collisions, respectively.} 
		\label{tab:coll_time}
  \vspace{-0.4cm}
\end{minipage}
\end{figure}
}

%
%

    To evaluate our external force and torque estimation method, the hybrid vehicle was manually flown into obstacles at various orientations, as shown in Figure  \ref{tab:coll_time}. 
	The visible drift on the estimated vertical external force $F_e^z$ is caused by the draining battery and lack of rotors feedback. 
	As only the estimated orientation of the contact force is used for both CIO and our reactive planner, detecting a collision and estimating the orientation of $\mathbf{F}_e$ is sufficient to enable resilient navigation.  
	Using the collision detection method in \eqref{eq:collision_threshold}, all but one collision are successfully detected from forces with minimal tuning efforts, whereas wheel encoders are able to detect the collision at 43s. 
	By comparing the true collision direction to the estimation results, all estimated external forces detected without wheel encoders present a reasonable orientation estimation accuracy which can be used for CIO or the reactive planner, validating our approach.
%
%
%

\begin{figure}[!htb]
\begin{minipage}{1.0\textwidth}
		\centering
		\subfigure{
			\includegraphics[width=45mm]{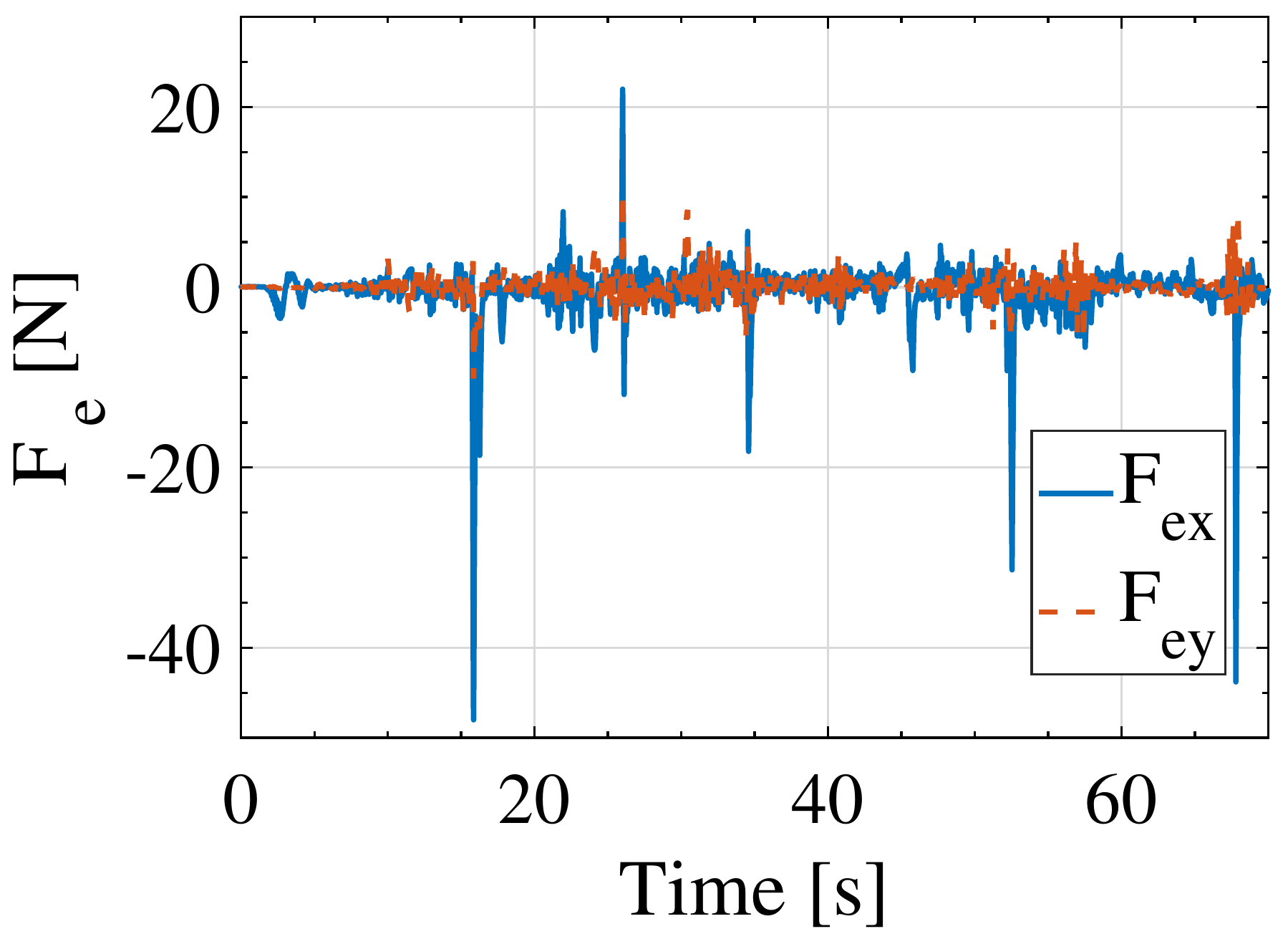}
			\label{fig:front_collision_wFrc_FeENC}}
		\subfigure{
			\includegraphics[width=45mm]{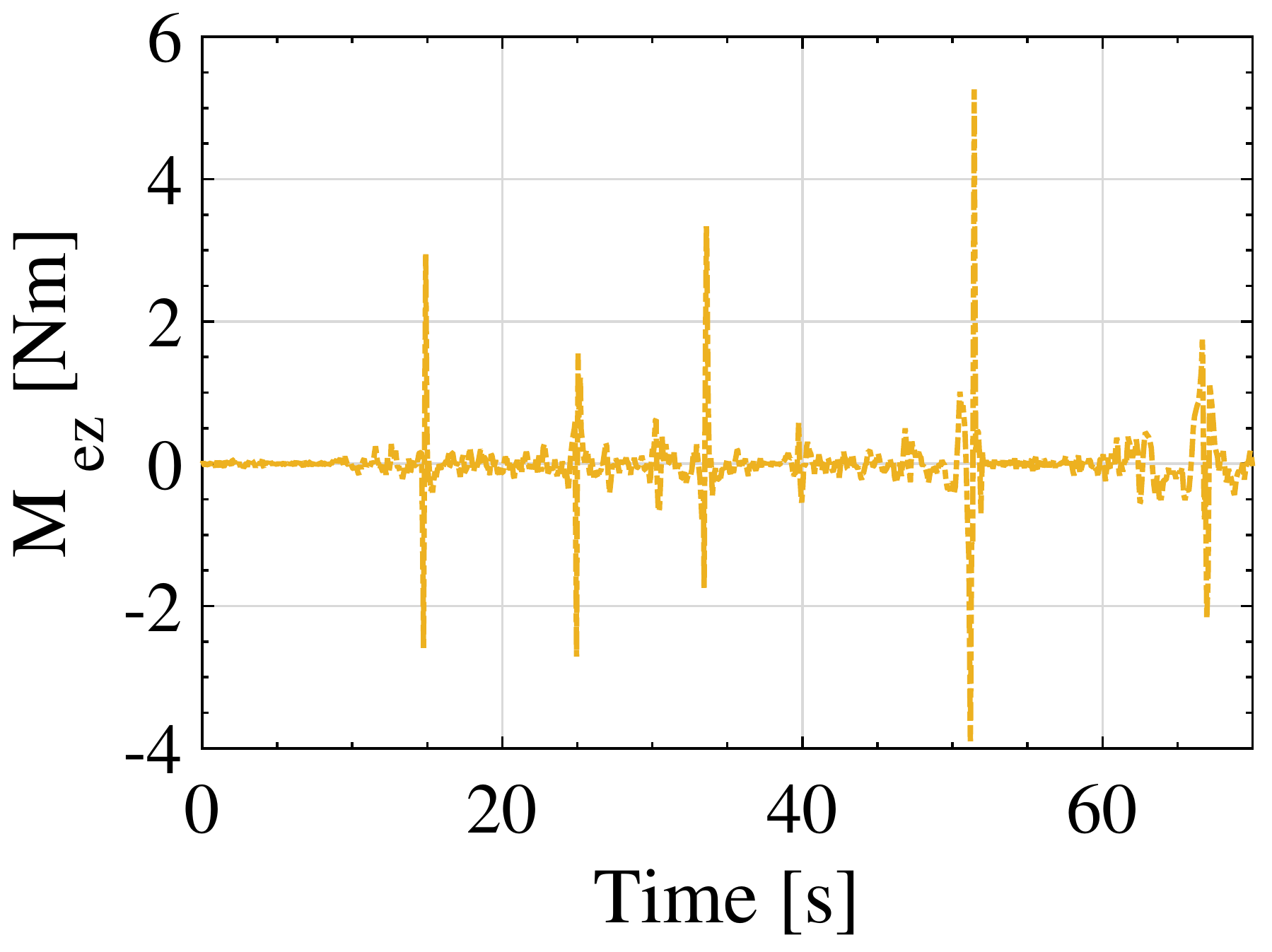}
			\label{fig:front_collision_wFrc_MezENC}}
\end{minipage}%
\par
\begin{minipage}{1.0\linewidth}
  \centering
	\begin{tabular}{|c||c|c|c|c|c|c|c|}
	\hline 
	time & 16s & 26s & 32s & 34s & 41s & 52s & 68s \\
	\hline
	direction & F (0$^\circ$)& F (0$^\circ$)& BR(-60$^\circ$)& F (0$^\circ$) &F (0$^\circ$) & F (0$^\circ$) & F (0$^\circ$) \\
			\hline 
			estimated &$3.9^\circ$& $-8.7^\circ$& $-73.9^\circ$& -21.0& - & $0.8^\circ$ & $-19.6^\circ$ \\
			\hline 	\end{tabular} 
	\caption{Force estimation (top) and ground truth of collision times  and directions  (bottom) for the rolling mode. The notation for contact directions is defined in Figure \ref{tab:coll_time}.}
	\label{tab:coll_time_roll}
\end{minipage}
\end{figure}

	Similarly, force estimation experiments while rolling were conducted, 
	as shown in Figure \ref{tab:coll_time_roll}. 
    Again, all collisions are correctly detected using both an IMU and wheel encoders and the direction of the forces estimated without wheel encoders are accurate. 
	
    \subsection{State Estimation using Contact Inertial Odometry}\label{sec:results:cio_est}
    We conduct experiments to validate our CIO algorithm and show that our method is able to correct for velocity estimation errors.  We compare our method against an estimate of the ground truth by fusing the measurements of the IMU with pose estimates of ORB-SLAM \cite{orb_slam} (i.e., a monocular simultaneous localization and mapping algorithm) running on RealSense RGBD data.  To demonstrate that our approach is useful for both aerial and hybrid vehicles, we first show results for  flying where the robot collides laterally with an obstacle. Then, we show that bouncing against the ground while flying can improve state estimation as well. We do not include experiments for state estimation while rolling, since wheel encoders would provide better velocity estimates than our method and this is already explored in the literature. 
    All flight experiments are performed using a hand-held safety tether, due to safety regulations.

    \subsubsection{Flying}
    Our CIO algorithm has been extensively tested in flight and is able to reliably correct for velocity drift. In Figure \ref{fig:front_vel_update}, we show a typical collision with an obstacle. In such cases, the estimated collision force can be used to (1) provide a parallel velocity measurement to update the state within an EKF as described in Section \ref{sec:cio} and (2) provide a reference direction for the reactive planner described in Section \ref{sec:reactive_planner}, which aims to avoid obstacles and continue exploration of the environment.  The information from a collision can be used as a measurement to successfully correct for IMU drift, when compared against an estimate of ground truth (ORB-SLAM + IMU fused with an EKF) as shown in Figure \ref{fig:front_vel_update}. In this experiment, vision-based state estimation is used for closed-loop control, whereas experiments in Figures \ref{fig:cio_up_down_bouncing} and \ref{fig:cio_in_action} use CIO only for state estimation, reactive planning and  control.
    \vspace{-7mm}
    
\newcommand{\cioXY}[1]{\includegraphics[clip, trim=0cm 1.1cm 0cm 1.1cm,  
    width=1.0\linewidth]{fig/seq_flying_top/cropped/#1}}
{
\begin{figure}[!htb]
\centering
\begin{minipage}{.28\textwidth}
  \centering
    \cioXY{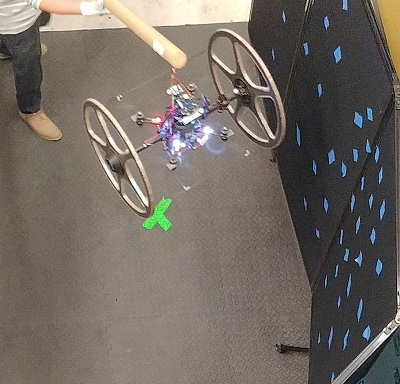}
\end{minipage}%
\begin{minipage}[t]{.06\textwidth}
  \centering  
  \hfill
\end{minipage}%
\begin{minipage}{0.28\textwidth}
  \centering
    \cioXY{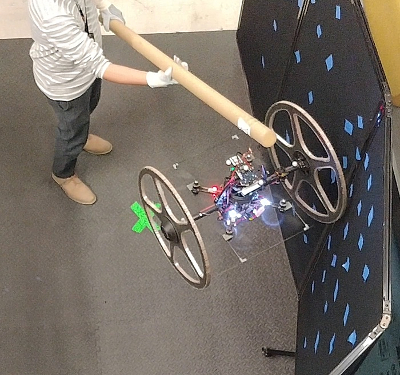}
\end{minipage}%
\begin{minipage}[t]{.06\textwidth}
  \centering  
  \hfill
\end{minipage}%
\begin{minipage}{0.28\textwidth}
  \centering
    \cioXY{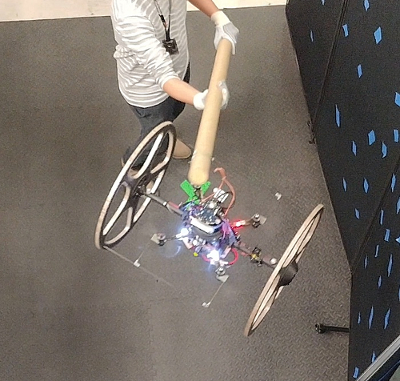}
\end{minipage}%
\par
\begin{minipage}{1.0\textwidth}
  \centering
  \vspace{0.5cm}
\end{minipage}%
\par
\vspace{-0.25cm}
\begin{minipage}{1.0\textwidth}
  \centering
	\includegraphics[width=0.95\linewidth]{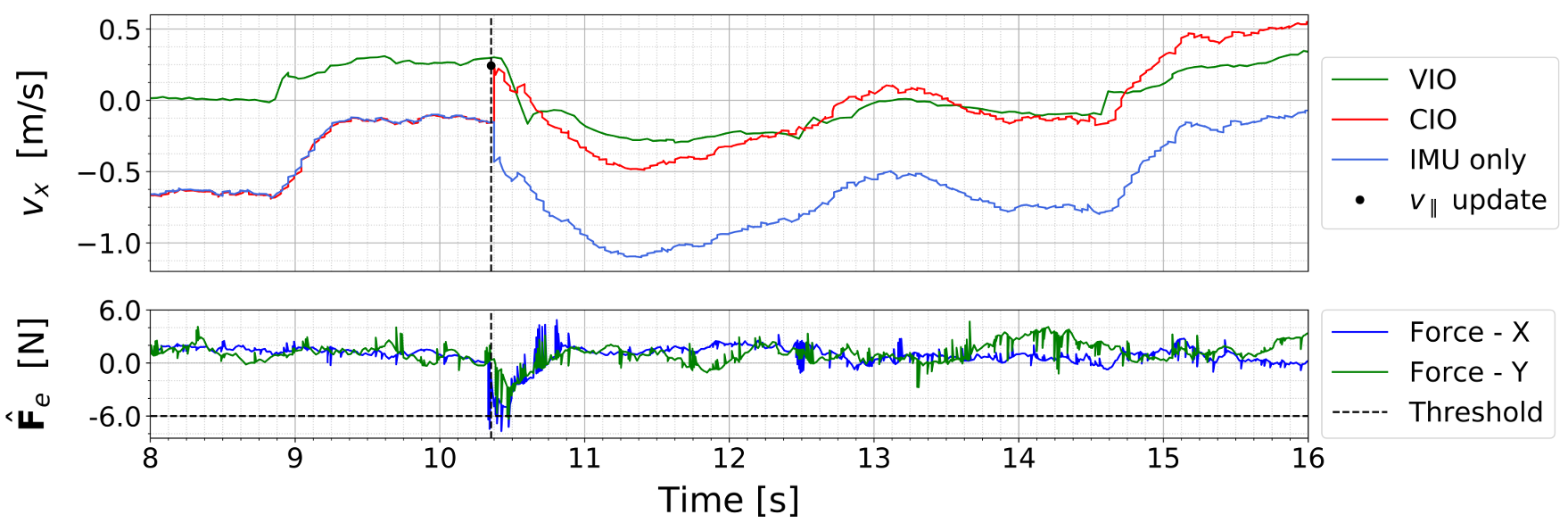}
	\vspace{-0.2cm}
	\caption{Parallel velocity measurement update for a frontal collision, correcting for the velocity estimate error.
	}
	\label{fig:front_vel_update}
\end{minipage} 
\end{figure}
}

	\subsubsection{Bouncing}
	To demonstrate the use of CIO for hybrid vehicles, we conduct an experiment where the robot is commanded to follow a vertical reference velocity $\mathbf{v}^{ref}$ alternating between up and down (last flight from the supplementary video).  
	In Figure \ref{fig:cio_up_down_bouncing} (top), we show a single flying and bouncing sequence performed by our hybrid platform.  
	Clearly, it is beneficial to periodically make contact with both the ground and lateral obstacles
	to obtain pseudo-measurements of 
	the velocity in all directions (bottom). 
    ORB-SLAM is used as ground truth for velocities and a height sensor is used to show the true distance to the ground and illustrate the up-down behavior. 
	
\newcommand{\verticalBouncing}[1]{\includegraphics[width=0.8\linewidth]{fig/maze/bouncing_updown/#1}}

{
\begin{figure}[!htb]
\vspace{-1.em}
\centering
\begin{minipage}{.22\textwidth}
  \centering
    \verticalBouncing{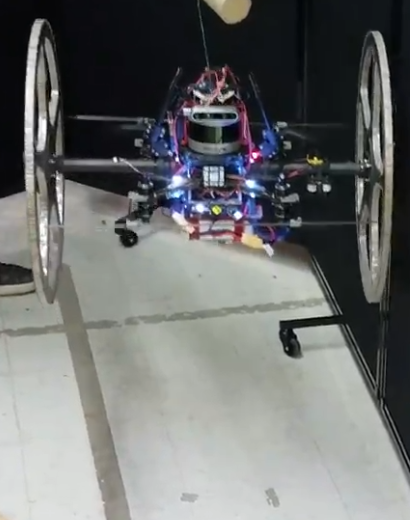}
\end{minipage}%
\begin{minipage}[t]{.04\linewidth}
  \centering 
  \hfill
\end{minipage}%
\begin{minipage}{.22\textwidth}
  \centering
    \verticalBouncing{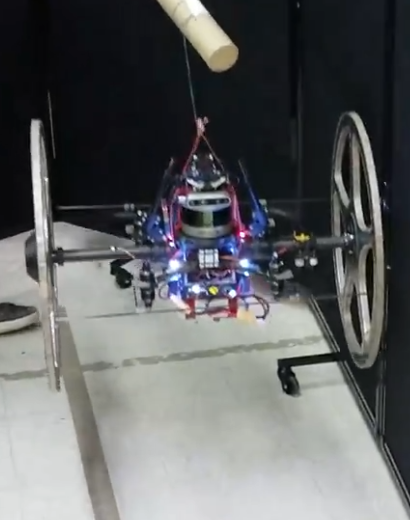}
\end{minipage}%
\begin{minipage}[t]{.04\linewidth}
  \centering 
  \hfill
\end{minipage}%
\begin{minipage}{0.22\textwidth}
  \centering
    \verticalBouncing{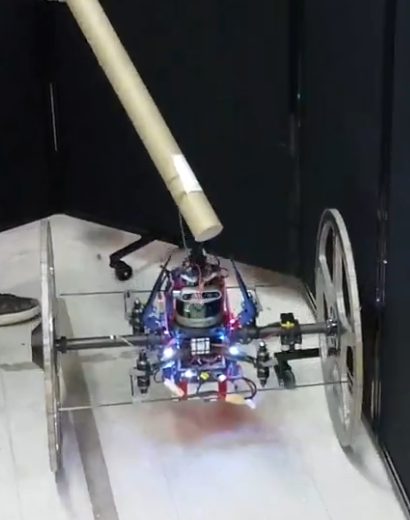}
\end{minipage}%
\begin{minipage}[t]{.04\linewidth}
  \centering 
  \hfill
\end{minipage}%
\begin{minipage}{0.22\textwidth}
  \centering
    \verticalBouncing{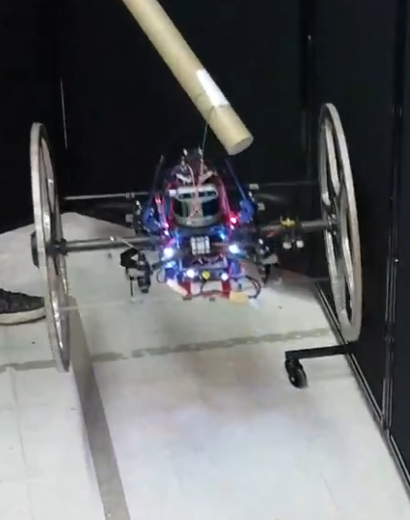}
\end{minipage}%
\par
\begin{minipage}{1.0\textwidth}
  \centering
  \vspace{0.25cm}
\end{minipage}%
\par
\begin{minipage}{1.0\textwidth}
  \centering
   \includegraphics[width=1.0\linewidth]{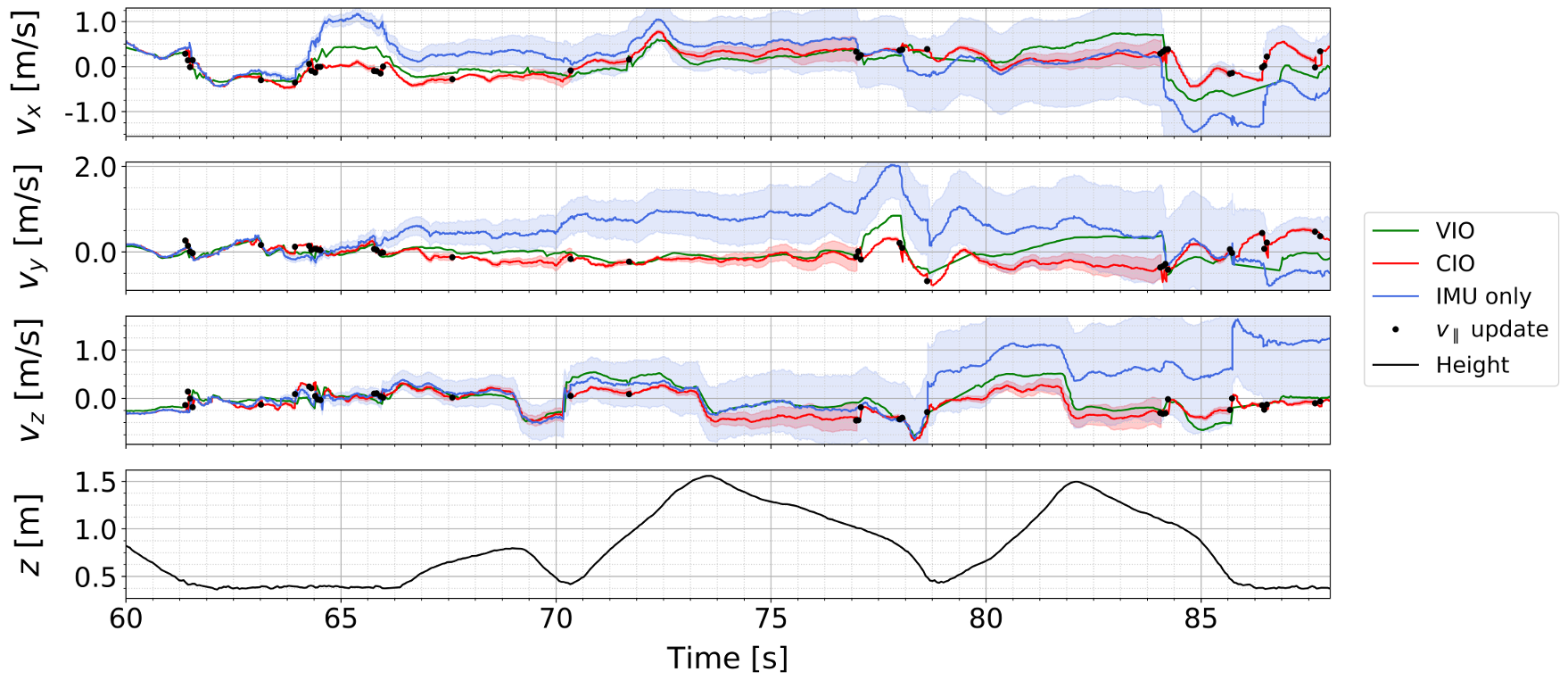}
  \caption{Flying and bouncing experiment with pseudo-measurements of the velocity.}
  \label{fig:cio_up_down_bouncing}
\end{minipage} 
\end{figure}
}
%
%
\vspace{-2em}
    \subsubsection{Discussion}
    Experimental results show
	that the velocity estimates produced by CIO using the parallel velocity defined in \eqref{eq:vel_parallel} are not highly accurate. 
	However, we stress that the goal of this paper is not to provide a highly accurate measurement model which can correct the estimates of the velocity to closely match the true velocity.  In fact, this is infeasible with this method since (1) the measurement model is a pseudo-measurement generated by assuming a direct measurement of the velocity, which we set to a value depending on the current estimate of the state and collision force, and (2) IMU acceleration estimates are corrupted by noise and drift, which are unobservable given our estimation method.  Correcting for such drift would require to hold contact for a prolonged amount of time to perform a zero-velocity update $\mathbf{z}_k = \mathbf{v}_k+\mathbf{w}_k = [0;0;0]$, 
	which we deem not acceptable, as the goal of this work is fast traversal of an environment when onboard exteroceptive sensors fail.
	This is in contrast to work where contacts are used to improve state estimates generated with exteroceptive sensors such as cameras \cite{nisar_vimo_2019},  whereas our goal is to fly without suffering catastrophic crashes using an IMU sensor only, requiring more drastic assumptions on the measurement model.

    Notice that for the sake of simplicity, we make the assumption that at a collision, the robot does not lose energy in the direction parallel to the wall.  To be more aligned with this assumption, we should tune the covariance of the pseudo measurement such that the variance remains constant along the axis of the parallel velocity. Although for the sake of simplicity we do not do this here, we expect that this should improve the performance for our filter.  
    

	\subsection{Autonomous Navigation Through a Maze}\label{sec:results:maze}
	
	We demonstrate that it is possible to traverse a cluttered environment by leveraging contact information.  To do so, we combine our contact detection and estimation method, CIO, with our reactive planner.  We constrain our hybrid vehicle to flight only and use the typical equations of the dynamical system of a quadrotor in \eqref{eq:dynamics_quadrotor} to show that our approach can be used on any type of quadrotor vehicle. 
	For safety reasons, for this experiment we use a  LiDAR height sensor in addition to the IMU sensor and perform feedback control on the height estimate from the height sensor.  However, we do not incorporate the height as a measurement update for the EKF.  This constrains the movement of the vehicle to a safe range around a desired height above the ground.  The robot is free to drift in the horizontal $x,y$ plane and collide with obstacles.
	
\begin{figure}[!htb]
      \centering
	\includegraphics[width=1.0\linewidth]{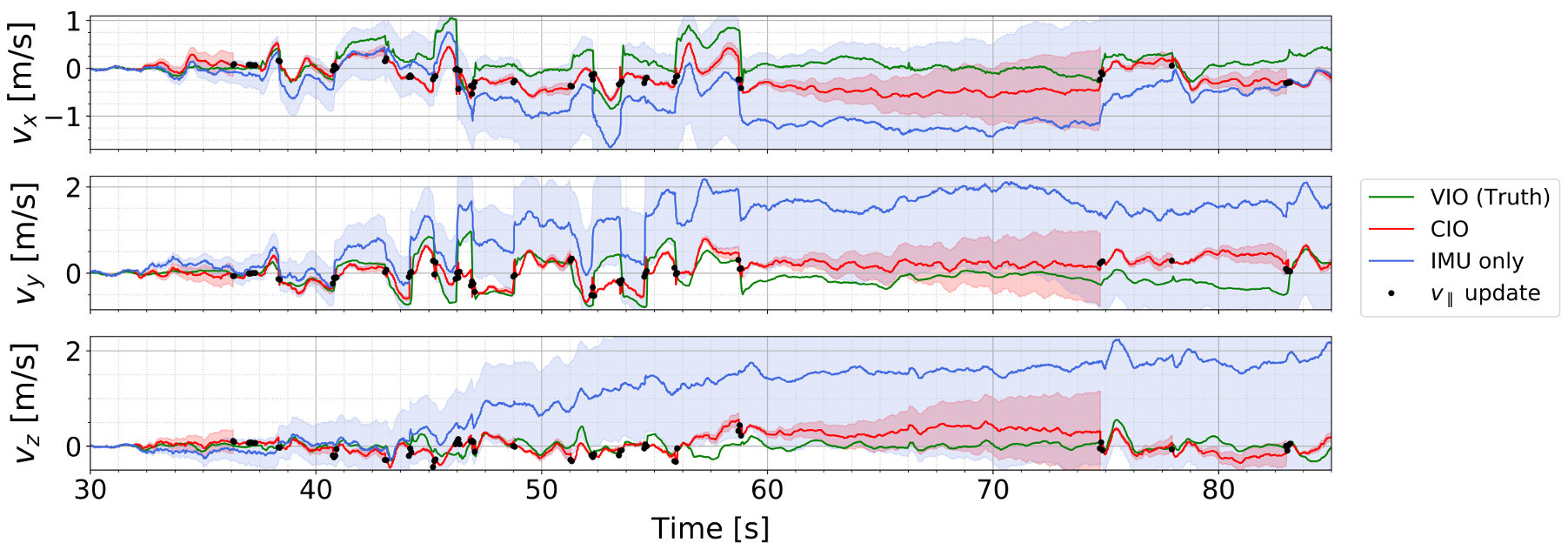}
      \vspace{-3mm}
      \caption{
      {\small Flying experiment in a maze, shown in Figure \ref{fig:blam_path}. Flying on IMU only causes the velocity estimates to drift, whereas contact information can be exploited as pseudo measurements of the velocity to correct for estimation error.}}
      \label{fig:cio_in_action}
      \vspace{-0.5cm}
    \end{figure}

	In Figure \ref{fig:blam_path}, we show a reconstruction of the maze environment from LiDAR point clouds.  The goal is for the robot to traverse this environment without perceiving it, since it flies using an IMU only.  At each collision, a parallel velocity update is performed and a new reference velocity is sent to the controller.  In Figure \ref{fig:cio_in_action}, we show that the velocity updates bound the velocity estimates, such that the robot is able to navigate autonomously without exteroceptive sensors.  
	
	We also plot the estimates from an EKF which does not use any collision updates and only uses the IMU for the prediction step (in blue).  Without collision updates, the vehicle would quickly accelerate due to feedback on drifted velocity estimates, and crash into obstacles at high speeds.   Furthermore, the vehicle would keep attempting to increase its speed as it pushes against the obstacle in contact, which is clear from the plotted standard deviations from the predictions of each EKF, with and without collision updates.  Without CIO, the EKF quickly diverges, which is expected as the state is unobservable.  In contrast, the CIO measurement updates constrain the drift in the velocity estimate and bounds its error.
	Therefore, our CIO algorithm and reactive planner are effective tools for navigating in these situations when no other sensors are available.  In the supplementary video, we demonstrate flying in the dark with all sensors obscured, and reliably achieve similar results.



	\section{Conclusion}\label{sec:conclusion}
	\vspace{-0.13cm}
	\vspace{-0.5em}
	In this work, we demonstrated autonomous navigation in an unknown cluttered environment using an IMU only. 
	To do so, we derived a pseudo-measurement model to update velocities by exploiting contact information, and designed a reactive planner encouraging exploration using estimated contact forces. 
	To fully demonstrate the capabilities of our approach, we 
	developed a force estimation method for a hybrid vehicle,  and validated each algorithmic component through hardware experiments in both flying and rolling modes. 
	This work can be used as a safety fallback on drones to recover from VIO failures or on micro-drones that do not have the payload capacity to carry sensors such as LiDARs, cameras, etc. or powerful computers to process their data.
	 

    Future work will include perception-aware planning to encourage collisions in optimal directions and frequency to maintain an acceptable state estimation error \cite{Slap_2018,Firm_2014}. 
    %
    Also, the accuracy of the CIO algorithm could be improved, by (1) investigating the use of error-state EKFs and other promising formulations \cite{Santamaria-NavarroAuro18,invariant_bonnabel}, (2) analyzing the rigid body kinematics to derive more informative measurement updates, as is done in legged robotics, (3) modeling the properties of contact surfaces and the loss of energy during collisions and deriving a measurement model using a high frequency IMU, and (4) varying and tuning the covariance of the pseudo-measurement update and contact detection threshold, e.g., using machine learning methods. 
    %
    Furthermore, active collision-based localization in a prior map in the absence of exteroceptive sensors could be a promising research direction.  
    This would allow the robot to perform behaviors such as returning to its original starting point using an IMU only.
    Finally, we welcome future research to formalize the idea of exploiting collisions instead of treating them as constraints which would allow to expand the  set of safe states for planning and control.


\vspace{-0.13cm}
	\section*{Acknowledgement}
\vspace{-1em}
{\small 
	This research was carried out at the Jet Propulsion Laboratory, California Institute of Technology, under a contract with the National Aeronautics and Space Administration.
    T. Lew is partially supported by the 242 Program of the Hubert Tuor Foundation. Tomoki Emmei is supported by JSPS KAKENHI Grant Number 18J14169.
    The authors 
    thank Dr. Matthew J. Anderson, Leon Kim and 
    members of the CoSTAR 
    team for their incredible support with experiments and hardware. 
	}

\renewcommand{\baselinestretch}{0.89} 
\bibliographystyle{styles/bibtex/splncs_srt}
\vspace{-0.5em}
\bibliography{ref}

\appendix
\section{Contact Position Estimation}\label{apdx:contact_point_position}
\subsection{Analytical Solution}
This section presents the analytical solution of the contact point estimation method presented in Section \ref{sec:external_force_est}. 
For conciseness, we denote $\tilde{F}_e^z:=(F_e^z-m_t g)$. 
%
The positions of the estimated contact forces on the left and right wheels are decomposed into two sets of equations depending on whether $F_y<0$ or $F_y\geq 0$.
\\

First, if $F_y < 0$, the solution for the position of the contact point on the left wheel is given as
\begin{subequations}
\begin{align}
        p^l_{x} =& \frac{-a_1c_1 -\mathrm{sgn}(F_e^x)|b_1|\sqrt{R^2(a_1^2+b_1^2)-c_1^2}}{a_1^2+b_1^2}\\
        p^l_{z} =& \frac{-a_1c_1 - \mathrm{sgn}(\tilde{F}_e^z)|a_1|\sqrt{R^2(a_1^2+b_1^2) -c^2}}{a_1^2+b_1^2},
        \end{align}
\end{subequations}
with $a_1 = M_{e}^x + L\tilde{F}_e^z$, $b_1 = M_{e}^z - LF_{e}^x$ and $c_1 = 2LM^l_w$. Using these equations, the estimated contact point on the right wheel is derived as  
\begin{subequations}
        \begin{align}
        p^r_{x} =& \frac{-A_1C_1 - \mathrm{sgn}(F_e^x)|B_1|\sqrt{r^2(A_1^2+B_1^2) -C_1^2}}{A_1^2+B_1^2}\\
        p^r_{z} =& \frac{-A_1C_1 - \mathrm{sgn}(\tilde{F}_e^z)|A_1|\sqrt{R^2(A_1^2+B_1^2)-C_1^2}}{A_1^2+B_1^2},
\end{align}
\end{subequations}
with $A_1 = (- L^2\tilde{F}_e^{z2} + F_{e}^yp^l_zL\tilde{F}_e^z + M_{e}^{x2} + F_{e}^yp^l_zM_{e}^x)$, $B_1 = M_{e}^xM_{e}^z + L^2F_{e}^x\tilde{F}_e^z + LM_{e}^xF_{e}^x + LM_{e}^z\tilde{F}_e^z + 2LM_{\rm w}^lF_{e}^y + M_{e}^zF_{e}^yp^l_z - LF_{e}^xF_{e}^yp^l_z$ and $C_1 = - 2LM_{e}^xM_{\rm w}^r - 2L^2M_{\rm w}^r\tilde{F}_e^z$.
\\

On the other hand, if $F_e^y \geq  0$, the position of the contact point on the  left and right wheels are computed as
\begin{subequations}
\begin{align}
        p^r_{x} =& \frac{-a_2c_2 -\mathrm{sgn}(F_e^x)|b_2|\sqrt{R^2(a_2^2+b_2^2)-c_2^2}}{a_2^2+b_2^2}\\
        p^r_{z} =& \frac{-a_2c_2 - \mathrm{sgn}(\tilde{F}_e^z)|a_2|\sqrt{R^2(a_2^2+b_2^2) -c^2}}{a^2+b^2}\\
        p^l_{x} =& \frac{-A_2C_2 - \mathrm{sgn}(F_e^x)|B_2|\sqrt{R^2(A_2^2+B_2^2) -C_2^2}}{A_2^2+B_2^2}\\
        p^l_{z} =& \frac{-A_2C_2 - \mathrm{sgn}(\tilde{F}_e^z)|A_2|\sqrt{R^2(A_2^2+B_2^2)-C_2^2}}{A_2^2+B_2^2},
\end{align}
\end{subequations}
with  $a_2 = M_{e}^x - L\tilde{F}_e^z$, $b_2 = M_{e}^z + LF_{e}^x$, $c_2 = -2LM^r_w$, $A_2 = (M_{e}^x - L\tilde{F}_e^z)(M_{e}^x + L\tilde{F}_e^z + F_{e}^yp^r_z)$, 
$B_2 = M_{e}^xM_{e}^z + L^2F_{e}^x\tilde{F}_e^z - LM_{e}^xF_{e}^x - LM_{e}^z\tilde{F}_e^z - 2LM_{\rm w}^rF_{e}^y + M_{e}^zF_{e}^yp^r_z + LF_{e}^xF_{e}^yp^r_z$ and 
$C_2 = 2LM_{e}^xM_{\rm w}^l - 2L^2M_{\rm w}^l\tilde{F}_e^z$
.

	\subsection{Results: Contact Point Estimation}
\begin{figure}[!htb]
		\centering
		\includegraphics[width=120mm]{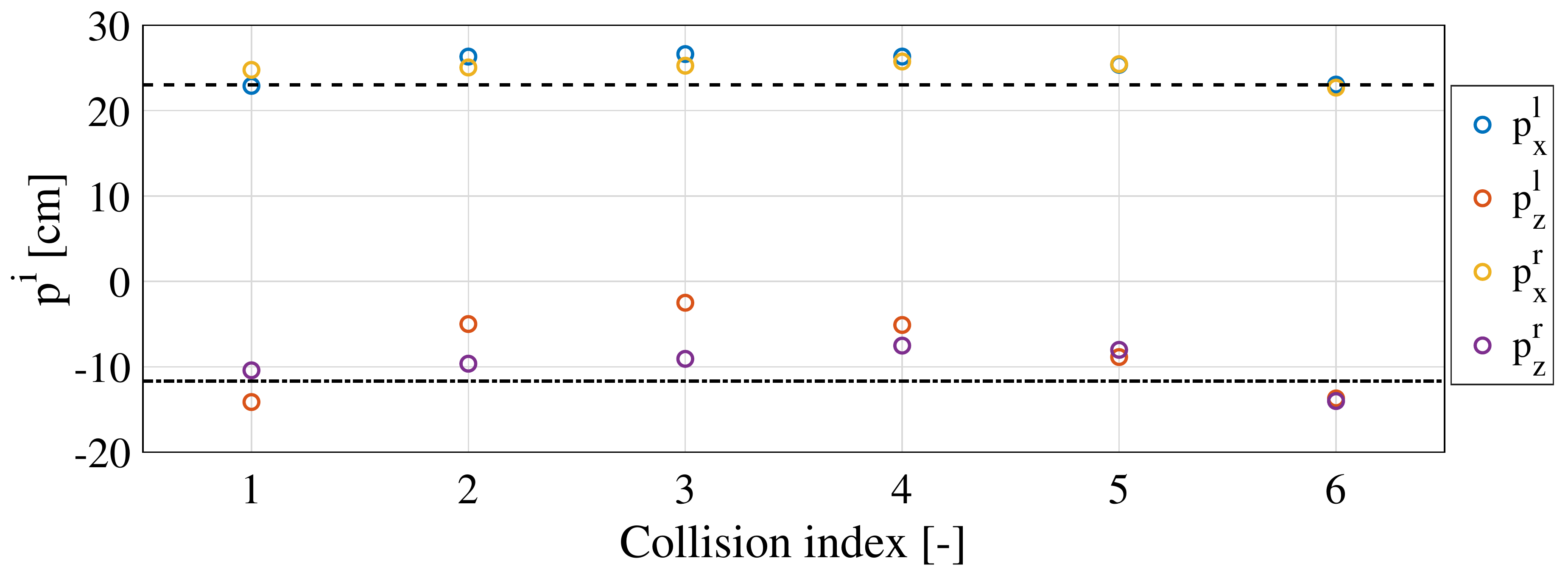}
		\caption{Estimated contact point positions in rolling mode, with ground truth indicated by straight lines. The height of the contact point  ($\mathrm{p_z^l},\mathrm{p_z^r}$) is negative since the origin is set to the center of the robot wheels. Each data point corresponds to a different experiment.}
		\vspace{-0.5cm}
		\label{fig:contact_exp}
	\end{figure}

	Experiments are conducted to validate our proposed contact point estimation method. The results are shown in Figure \ref{fig:contact_exp}.
	
	In each experiment, the hybrid vehicle Rollocopter is driven on a flat ground and collided frontally with a fixed box-shaped obstacle with a height of 15cm. From the known dimensions of the box and of the wheels, it is possible to determine the true contact point position. 
	As shown in Figure \ref{fig:contact_exp}, the estimated positions $p^i_x$ and $p^i_z$ are close to the their true value for each one of the 6 experiments, validating our approach. 
	This method could be used in future work to determine whether the contact point is caused by a collision in front of or behind the vehicle, or if it is detected due to rough terrain, providing additional useful information, e.g. for path planning.
    

\newcommand{\PLH}{{\mkern-3mu\times\mkern-3mu}}
\newcommand{\PLUS}{{\mkern-3mu{+}\mkern-3mu}}
\newcommand{\MINUS}{{\mkern-3mu{-}\mkern-3mu}}

\section{Derivation of Nonholonomic Model}
This section derives additional nonholonomic constraints for the rolling mode of the hybrid vehicle.
\subsection{Derivation of Nonholonomic Constraint}\label{apdx:sec:nonholonomic_constraints}
To derive \eqref{eq:NonHolo}, we first express the velocities at the contact points on the ground as
{\small
\begin{subequations}\label{eqs:velocity_contact_points}
\begin{align}
      \bm{v_c^l} &=  \begin{pmatrix}
      v_x \\ 
      v_y \\ 
      0
      \end{pmatrix} \PLUS
      \begin{pmatrix}
      0\\ 
      0\\ 
      \omega_z
      \end{pmatrix}
		{\PLH}
		 \begin{pmatrix}
		 0 \\ 
		 L \\ 
		 0
		 \end{pmatrix} 
		 \PLUS
      \begin{pmatrix}
0\\ 
\omega_y {+} \gamma_l\\ 
0
\end{pmatrix}
{\PLH}
\begin{pmatrix}
0 \\ 
0 \\ 
-R
\end{pmatrix} 
{=}
      \begin{pmatrix}
		v_x {-} L\omega_z {-} R(\omega_y{+}\gamma_l)\\ 
		v_y\\ 
		0
		\end{pmatrix}
      \\ 
      \bm{v_c^r} &=  \begin{pmatrix}
v_x \\ 
v_y \\ 
0
\end{pmatrix} \PLUS
\begin{pmatrix}
0\\ 
0\\ 
\omega_z
\end{pmatrix}   
{\PLH}    
\begin{pmatrix}
0 \\ 
-L \\ 
0
\end{pmatrix} 
\PLUS
      \begin{pmatrix}
0\\ 
\omega_y \PLUS \gamma_r\\ 
0
\end{pmatrix}
{\PLH}
\begin{pmatrix}
0 \\ 
0 \\ 
-R
\end{pmatrix}
{=}
\begin{pmatrix}
v_x {+} L\omega_z {-} R(\omega_y{+} \gamma_l)\\ 
v_y\\ 
0
\end{pmatrix}
.
\end{align}
\end{subequations}
}%
Assuming that the wheels of the hybrid vehicle remain in contact with the ground and that no slip occurs, 
As long as the wheels keep contact with the ground and no slip occurs, $\mathbf{v}_c^l=\mathbf{v}_c^r = \mathbf{0}$ hold. Therefore, \eqref{eqs:velocity_contact_points} can be equivalently expressed as
	\begin{align}
		v_x = \frac{R}{2}(\gamma_r + \gamma_l+2\omega_y),\ \ 
		v_y = 0,\ \ 
		\omega_z = \frac{R}{2L}(\gamma_r - \gamma_l)
		,
	\end{align}
which is equivalent to \eqref{eq:NonHolo}.

\subsection{Derivation of Force Estimation for Rolling Mode}\label{apdx:sec:force_est_rolling_mode}
In this section, we show how to take into account nonholonomic constraints in our force estimation method.  
First, the nonholonomic motion of the hybrid vehicle implies that
\begin{equation}
    v_y,v_z,\dot{v}_y,\omega_x = 0
    .
\end{equation}

Also, by including the left and right wheel rolling resistance forces $F_d^l,F_d^r$ acting in the rolling direction, 
 the dynamics of the hybrid vehicle in \eqref{eq:dynamics_flying} are rewritten as
\begin{subequations}
\begin{align}
	m_t\dot{v}_x&=F_{in}^x+F_{e}^x-F_{d}^l - F_{d}^r \label{eq:nonholoX}\\
	m_tv_x\omega_z &=F_{e}^y \label{eq:nonholoY}\\
    (I_t^z+2m_wL^2)\dot{\omega}_z &=(M_{in}^z+M_{e}^z-L(F_{d}^r-F_{d}^l)). \label{eq:nonholoGamma}
\end{align}
\end{subequations}

Furthermore, by analyzing the dynamics of each wheel in Equation  \eq{mdl_wheel}, we have
\begin{equation}
    I_{\rm w}(\dot{\gamma}_i + \dot{\omega}_y)= RF_{d}^i. \label{eq:moments_Fd}
\end{equation}

Therefore, using the nonholonomic constraint \eqref{eq:NonHolo} to replace $v_x,\dot{v}_x,\omega_z,\dot\omega_z$ and the previous result to replace $F_d^i$, the equations above are rewritten as
\begin{subequations}
\begin{align}
	\left(\frac{m_tR}{2} + \frac{2I_{\rm w}}{R}\right)(\dot{\gamma}_r+\dot{\gamma}_l+2\dot{\omega_y}) =F_{in}^x + F_{e}^x \label{eq:nonholoX2}\\
	\frac{m_tR^2}{4L}(\gamma_r-\gamma_l)(\gamma_r+\gamma_l+2\omega_y)=F_{e}^y. \label{eq:nonholoY2}
\end{align}
\end{subequations}

Also,  \eqref{eq:nonholoGamma} can be rewritten by replacing $F_d^l,F_d^r,\dot\omega_z$ using Equations \eqref{eq:moments_Fd} and \eqref{eq:NonHolo} as
\begin{equation}
    \left(\frac{R}{2L}I_t^z+m_wLR+\frac{I_{\rm w}L}{R}\right)(\dot{\gamma}_r - \dot{\gamma}_l)= M_{in}^z+ M_{e}^z 
    .
\end{equation}

Finally, the external wrench $\{F_{e}^x ,F_{e}^y, M_{e}^z \}$ can be computed as
\begin{subequations}
\begin{align}
F_{e}^x &= \left(\frac{m_tR}{2} + \frac{2I_{\rm w}}{R}\right)(\dot{\gamma}_r + \dot{\gamma}_l +2\dot{\omega}_y) -F_{in}^x\\
F_{e}^y &= \frac{m_tR^2}{4L}(\gamma_r-\gamma_l)(\gamma_r+\gamma_l+2\omega_y)\\
M_{e}^z &= 	\left(\frac{R}{2L}I_t^z+m_wLR+\frac{I_{\rm w}L}{R}\right)(\dot{\gamma}_r - \dot{\gamma}_l)-M_{in}^z
.
\end{align}
\end{subequations}

\end{document}